\RequirePackage{fix-cm}
\documentclass[5p,sort&compress]{elsarticle}  
\usepackage{graphicx}
\journal{}
\usepackage[numbers]{natbib}
\usepackage{graphicx}
\usepackage{amsmath,amssymb,amsfonts}
\usepackage{amsthm}
\usepackage{algorithmic}
\usepackage{color}
\usepackage[toc,page]{appendix}
\usepackage{environ}
\usepackage{array}
\usepackage{pdflscape}

\usepackage{tikz}
\usetikzlibrary{shapes.arrows}

\usepackage{natbib}
\usepackage{color}
\usepackage{rotate}
\usepackage{color}
\usepackage{soul}
\usepackage{listings}
\usepackage{hyperref}

\newcommand{\bea}{\begin{eqnarray*}}
	\newcommand{\eea}{\end{eqnarray*}}
\newcommand{\bne}{\begin{equation*}}

\newcommand{\ede}{\end{equation*}}

\newcommand{\bnen}{\begin{equation}}
\newcommand{\eden}{\end{equation}}
\newcommand{\bean}{\begin{eqnarray}}
\newcommand{\eean}{\end{eqnarray}}
\newcommand{\bnsn}{\begin{subequations}}
	\newcommand{\edsn}{\end{subequations}}

\newcommand{\bna}{\begin{array}}
	\newcommand{\eda}{\end{array}}
\newcommand{\bnm}{\begin{enumerate}}
	\newcommand{\edm}{\end{enumerate}}
\newcommand{\bni}{\begin{itemize}}
	\newcommand{\edi}{\end{itemize}}

\newtheorem{thm}{Definition}
\newtheorem{rem}{Remark}

\begin{document}
	
	\begin{frontmatter}
		
		\title{Squashing activation functions in benchmark tests: towards eXplainable Artificial Intelligence using continuous-valued logic %and MCDM%for better interpretability of neural networks
		}
		\cortext[mycorrespondingauthor]{Corresponding author}
		
		\author[1]{Daniel Zeltner}
		\ead{daniel.zeltner@outlook.com}
		
		\author[1]{Benedikt Schmid}
		\ead{be.p.schmid@gmx.de}
		
		\author[2,3]{G\'abor Csisz\'ar}
		\ead{Gabor.Csiszar@mp.imw.uni-stuttgart.de}
		
		\author[1,3]{Orsolya Csisz\'ar\corref{mycorrespondingauthor}}
		\ead{csiszar.orsolya@nik.uni-obuda.hu}
		
		\address[1]{Faculty of Basic Sciences, University of Applied Sciences Esslingen, Esslingen, Germany}
		\address[3]{Physiological Controls Research Center, \'Obuda University, Budapest, Hungary}
		\address[2]{Institute of Materials Physics, University of Stuttgart, Stuttgart, Germany}

		\begin{abstract}
			%An important characteristic of neural networks is their interpretability. 
			Over the past few years, deep neural networks have shown excellent results in multiple tasks, however, there is still an increasing need to address the problem of interpretability to improve model transparency, performance, and safety. Achieving eXplainable Artificial Intelligence (XAI) by combining neural networks with continuous logic and multi-criteria decision-making tools is one of the most promising ways to approach this problem: by this combination, the black-box nature of neural models can be reduced. The continuous logic-based neural model uses so-called Squashing activation functions, a  parametric family of functions that satisfy natural invariance requirements and contain rectified linear units as a particular case. 
			This work demonstrates the first benchmark tests that measure the performance of Squashing functions in neural networks. 
			Three experiments were carried out to examine their usability and a comparison with the most popular activation functions was made for five different network types. The performance was determined by measuring the accuracy, loss and time per epoch. These experiments and the conducted benchmarks have proven that the use of Squashing functions is possible and similar in performance to conventional activation functions. Moreover, a further experiment was conducted by implementing nilpotent logical gates to demonstrate how simple classification tasks can be solved successfully and with high performance. 
			The results indicate that due to the embedded nilpotent logical operators and the differentiability of the Squashing function, it is possible to solve classification problems, where other commonly used activation functions fail.			
		\end{abstract}
		
	\end{frontmatter}
	
	\section{Introduction}
	While AI techniques, especially deep learning techniques, are revolutionizing the business and technology world, there is an increasing need
	to address the problem of interpretability and to improve model transparency, performance, and safety: a problem that is of vital importance to all our research community. This challenge is closely related to the fact that although deep neural networks have achieved impressive experimental results, especially in image classification, they have shown to be surprisingly unstable when it comes to adversarial perturbations: minimal changes to the input image may cause the network to misclassify it. Moreover, although machine learning algorithms are capable of learning from a set of data and of producing a model that can be used to solve different problems, the values of the accuracy or the prediction error are not enough, since these numbers only provide an incomplete description of most real-world problems. The interpretability of a machine learning model gives insight on its internal functionality to explain the reasons why it suggests making certain decisions. %Interpretability, in the context of machine learning, is defined as the ability to explain or present to a person in a comprehensive way. This means that a person must be able to understand why a model determines how values are set.
	%Depending on the situation, we may or may not care why the model suggested taking a specific decision. For example, i
	In low-risk environments, such as film recommendation, only the predictive performance of the model counts. However, in high-risk environments, such as the health care or the insurance sector, it is important to be able to explain why a decision was made. %Nowadays we have highly regulated industries such as the health care or the insurance sector and in their cases, the interpretability of neural networks is very important. A model must provide an
	%explanation of how it came to its result. 
	In this case, we need to have some reasonable explanations behind our decisions to be more convincing and also to avoid lawsuits claiming race-based, gender-based, or age-based bias \cite{XAI}. Understandability means that we are able to describe the computations by using words from natural human language. One of the main challenges here is that natural language is often imprecise (fuzzy), making it difficult to find the relation between imprecise words and mathematical algorithms.  This experience led to the design on fuzzy logic by Zadeh; see, e.g., \cite{belo, klir, mendel, ngu, zadeh}. The reason, why human-led control often leads to much better results than even the optimal automatic control is that humans use additional knowledge.
	
	The basic idea of continuous logic is the replacement of the space of truth values $\{T,F\}$ by a compact interval such as $[0, 1]$.  This means that the inputs and the outputs of the extended logical gates are real numbers of the unit interval, representing truth values of inequalities. Quantifiers $\forall x$  and $\exists x$ are replaced by $\sup_x$ and $\inf_x$, and logical connectives are continuous functions. Based on this idea, human thinking and natural language can be modeled in a more sophisticated way. 
	Among other families of fuzzy logics, nilpotent fuzzy logic is beneficial from several perspectives. The fulfillment of the law of contradiction and the excluded middle, and the coincidence of the residual and the S-implication \cite{Dubois,Trillasimpl} make the application of nilpotent operators in logical systems promising.  In \cite{iwobi,bounded,boundedimpl,boundedeq,aggr,ijcci}, an abundant asset of operators was examined thoroughly: in \cite{bounded}, negations, conjunctions and disjunctions, in \cite{boundedimpl} implications, and in \cite{boundedeq} equivalence operators. In \cite{aggr}, the aggregative operators were studied and a parametric form of a general operator $o_{\nu}$ was given by using a shifting transformation of the generator function.   Varying the parameters, nilpotent conjunctive, disjunctive, aggregative (where a high input can compensate for a lower one) and negation operators can all be obtained. It was also demonstrated how the nilpotent generated operator can be applied for preference modeling. Moreover, as shown in \cite{ijcci}, membership functions, which play a substantial role in the overall performance of fuzzy representation, can also be defined using a generator function. %In \cite{interpret}, the authors showed that in the field of continuous logic, nilpotent logical systems are the most suitable for neural computation. 
	
	In \cite{interpret, pref}, the authors introduced an idea of achieving eXplainable Artificial Intelligence (XAI) by combining neural networks with nilpotent fuzzy logic as a promising way to approach the problem: by this combination, the black-box nature of neural models can be reduced, and the neural network-based models can become more interpretable, transparent, and safe. In \cite{interpret}, the authors showed that in the field of continuous logic, nilpotent logical systems are the most suitable for neural computation. To achieve the transparency using logical operators, it is desirable to chose an activation function that fits the theoretical background the best. In the formulae of the nilpotent operators, the cutting function (Heaviside or binary threshold function) plays a crucial role. Although piecewise linear functions are easy to handle, there are areas where the parameters are learned by a gradient-based optimization method. In this case, the lack of continuous derivatives makes the application impossible. 
	To address this problem, a continuously differentiable approximation of the cutting function, the Squashing function, introduced in \cite{Gera}, was used in the nilpotent neural model \cite{interpret}, \cite{pref}. In \cite{why}, the authors explain the empirical success of Squashing functions by showing that the formulas describing this family (that contain rectified linear units as a particular case) follow from natural invariance requirements.

	This study provides the first benchmark tests that measure the performance of the Squashing functions in neural networks and also demonstrates the first steps towards the implementation of nilpotent logical gates.
	The article is organized as follows. % the second part describes the implementation of the continuous logic using this function.
	After recalling the most important preliminaries in Section \ref{Prel}, %Section \ref{sq} focuses on the definition and main properties of the squashing functions. 
	Section \ref{S1} provides three experiments to demonstrate the usability of squasing activation functions together with a comparison with the most popular activation functions for five different network types. The performance of these functions was determined by measuring accuracy, loss and time per epoch. These experiments and the conducted benchmarks have proven that the use of the Squashing function is possible and similar in performance to the conventional activation functions. 
	In Section \ref{S2}, a further experiment was conducted by implementing nilpotent logical gates to demonstrate how 
	%In the second part, the nilpotent neural networks were investigated in regards to the interpretability of neural networks. The key point presented is the implementation and validation of the continuous logic. The research shows that nilpotent logical operators can easily be implemented. 
	simple classification tasks can be performed successfully and with high performance. Due to their low complexity, these networks are easy to interpret and analyze. The results indicate that due to nilpotent logical operators and the differentiability of the Squashing function, it is possible to solve classification problems, where other commonly used activation functions fail.
	Finally, in Section \ref{Concl}, the main results are summarized.
	
	\section{Preliminaries}\label{Prel}
	
	First, we recall some important preliminaries regarding nilpotent logical systems and Squashing functions.
	%Among other families of fuzzy logics, nilpotent fuzzy logic is favourable from several perspectives. The fulfillment of the law of contradiction and the excluded middle, and the coincidence of the residual and the S-implication \cite{Dubois,Trillasimpl} make the application of nilpotent operators in logical systems promising.  In \cite{iwobi,bounded,boundedimpl,boundedeq,aggr,ijcci}, an abundant asset of operators was examined thoroughly: in \cite{bounded}, negations, conjunctions and disjunctions, in \cite{boundedimpl} implications, and in \cite{boundedeq} equivalence operators. In \cite{aggr}, the aggregative operators were studied and a parametric form of a general operator $o_{\nu}$ was given by using a shifting transformation of the generator function.   Varying the parameters, nilpotent conjunctive, disjunctive, aggregative (where a high input can compensate for a lower one) and negation operators can all be obtained. It was also demonstrated how the nilpotent generated operator can be applied for preference modeling. Moreover, as shown in \cite{ijcci}, membership functions, which play a substantial role in the overall performance of fuzzy representation, can also be defined using a generator function. In \cite{interpret}, the authors showed that in the field of continuous logic, nilpotent logical systems are the most suitable for neural computation. 
	\subsection{Nilpotent logical systems}
	As mentioned in the Introduction, in the field of continuous logic, nilpotent logical systems are the most suitable for neural computation. For more details about nilpotent systems see \cite{ iwobi,bounded,boundedimpl,boundedeq,aggr,ijcci,interpret}.
	In \cite{aggr}, the authors examined a general parametric operator, $o_{\nu}(\underline{x})$, of nilpotent systems. 
	\begin{thm}\cite{aggr} Let $f:[0,1]\rightarrow [0,1]$ be an increasing bijection, $\nu\in[0,1]$, and $\underline{x}=(x_1,\dots, x_n),$ where $x_i\in[0,1]$ and let us define the general operator by
		\begin{equation} \label{o}
		\begin{split}
		o_{\nu}(\underline{x})=f^{-1}\left[\sum\limits_{i=1}^n \left(f(x_i)-f(\nu)\right)+f(\nu)\right]=
		\\
		=f^{-1}\left[\sum\limits_{i=1}^n f(x_i)-(n-1)f(\nu)\right].
		\end{split}
		\end{equation}
	\end{thm}
	
	\begin{rem} \label{cd} Note that the general operator for $\nu=1$ is conjunctive, for $\nu=0$ it is disjunctive and for $\nu=\nu_*=f^{-1}\left(\frac{1}{2}\right)$ it is self-dual.
	\end{rem} 
	As a benefit of using this general operator, a conjunction, a disjunction and an aggregative operator differ only in one parameter of the general operator in Equation \eqref{o}. Additionally, the parameter $\nu$ has the semantic meaning of the level of expectation: maximal for the conjunction, neutral for the aggregation, and minimal for the disjunction. 
	
	\begin{table*}[t]
		\caption{The most important two-variable operators $o_{\underline{w}} (\underline{x})$}\label{operators}
		\begin{center}
			\begin{tabular}{ | l | l | l | l | l | l | l |}
				\hline
				& $w_1$ & $w_2$ & $C$ & $o_{\underline{w}}(x,y)$&for $f(x)=x$ &Notation  \\ \hline \hline
				\multicolumn{7}{|l|}{LOGICAL OPERATORS}\\	\hline
				disjunction& $1$ & $1$ & $0$  & $f^{-1}[f(x)+f(y)]$  &$[x+y]$ & $d(x,y)$ \\ \hline
				conjunction& $1$ & $1$ & $-1$  & $f^{-1}[f(x)+f(y)-1]$  &$[x+y-1]$& $c(x,y)$ \\ \hline
				implication& $-1$ & $1$ & $1$  & $f^{-1}[f(y)-f(x)+1]$  &$[y-x+1]$& $i(x,y)$ \\ \hline
				\multicolumn{7}{|l|}{MULTI-CRITERIA DECISION TOOLS}\\	\hline
				arithmetic mean& $0.5$ & $0.5$ & $0$  & $f^{-1}\left[\frac{1}{2}\left(f(x)+f(y)\right)\right]$  &$\frac{1}{2}(x+y)$ & $m(x,y)$ \\ \hline
				preference & $-0.5$ & $0.5$ & $0.5$  & $f^{-1}\left[\frac{1}{2}\left(f(y)-f(x)+1\right)\right]$  & $\frac{1}{2}\left(y-x+1\right)$ &$p(x,y)$ \\ \hline
				aggregative operator & $1$ & $1$ & $-0.5$  & $f^{-1}\left[f(x)+f(y)-\frac{1}{2}\right]$  & $\left[x+y-\frac{1}{2}\right]$ & $a(x,y)$ \\ \hline
			\end{tabular}
		\end{center}
	\end{table*} 
	Next, let us recall the weighted form of the general operator:
	\begin{thm}\cite{aggr} Let $\underline{w}\in \mathbb{R}^n, w_i>0,%=(w_1,\dots,w_n)$ and $w_i>0$ be real parameters, 
		f:[0,1]\rightarrow [0,1]$ an increasing bijection with $\nu\in[0,1], \underline{x}=(x_1,\dots, x_n),$ where $x_i\in[0,1].$ The weighted general operator is defined by
		\begin{equation}\label{a}
		o_{\nu,\underline{w}}(\underline{x}):=f^{-1}\left[\sum\limits_{i=1}^n w_i (f(x_i)-f(\nu))+f(\nu)\right].
		\end{equation}
	\end{thm}
	
	Note that if the weight vector is normalized; i.e. for $\sum_{i=1}^{n} w_i=1,$ 
	\begin{equation}
	o_{\nu,\underline{w}}(\underline{x})=f^{-1} \left(\sum_{i=1}^{n} w_i f(x_i)\right).
	\end{equation}
	For future application, we introduce a threshold-based operator in the following way. 
	\begin{thm} \cite{aggr} Let $\underline{w}\in \mathbb{R}^n, w_i>0, \underline{x}=(x_1,...x_n)\in [0,1]^n$, $\underline{\nu}=(\nu_1,...\nu_n) \in [0,1]^n$and let $f:[0,1]\rightarrow[0,1]$ be a strictly increasing bijection. Let us define the threshold-based nilpotent operator by
		\begin{equation*}
		o_{\underline{\nu},\underline{w}}(\underline{x})=
		f^{-1} \left[\sum_{i=1}^{n} w_i \left(f(x_i)-f(\nu_i\right))+f(\nu)\right]=
		\end{equation*}
		\begin{equation}	=f^{-1} \left[\sum_{i=1}^{n} w_i f(x_i)+C\right],\label{nn}
		\end{equation}
	\end{thm} 
	where 
	\begin{equation}
	C = f(\nu)-\sum_{i=1}^{n} w_i f(\nu_i).
	\end{equation}
	
	\begin{rem}Note that the Equation in \eqref{nn} describes the perceptron model in neural computation. Here, the parameters all have semantic meanings as importance (weights), decision level and level of expectancy. Table \ref{operators} shows how the logical operators and some multi-criteria decision tools, like the preference operator, can be implemented in neural models. 
	\end{rem}
	The most commonly used operators for $n=2$ and for special values of $w_i$ and $ C$, also for $f(x)=x$, are listed in Table \ref{operators}. 
	
	\subsection{Squashing Function as a differentiable parametric app\-roximation of the Heavi\-side Function}\label{sq}
	As highlighted in the Introduction, in the formulae of the nilpotent operators, the cutting function plays a critical role  (see Table \ref{operators}).
	To address the problem of the lack of differentiability, the following approximation, the so-called Squashing function (introduced in \cite{Gera}) was used in the nilpotent neural model \cite{interpret}, \cite{pref} .

	\begin{figure}
		\centering        	\includegraphics[width=0.45\textwidth]{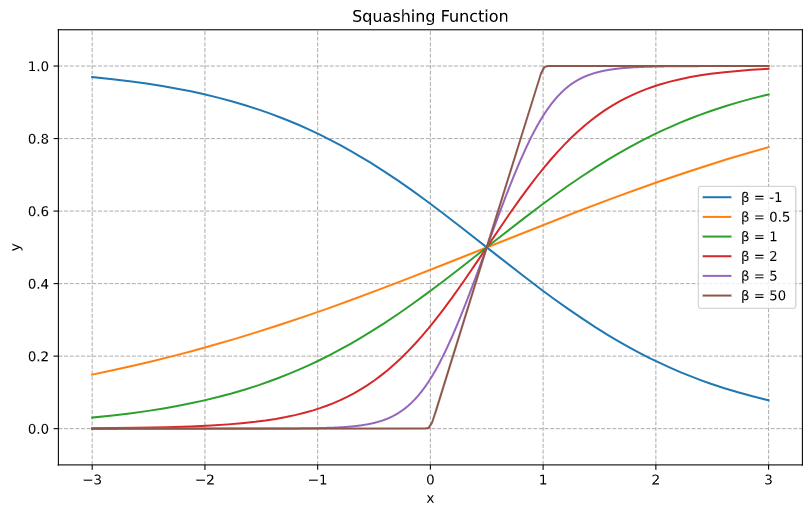}
		
		\caption{Squashing functions for $a=0.5$, $\lambda=1,$ for different $\beta$ values ($\beta_1=0.5,$ $\beta_2=1, $$\beta_2=2, $$\beta_3=5, $ and $\beta_4=50$) }\label{fig:squash}
	\end{figure}
	
	\begin{thm}
		The Squashing function \cite{ijcci,Gera} is defined as 
		\begin{equation}
		S^{(\beta)}_{a,\lambda}(x) = \frac1{\lambda\beta}\ln\frac{1+e^{\beta\left(x-(a-\lambda/2)\right)}}{1+e^{\beta\left(x-(a+\lambda/2)\right)}} = \frac1{\lambda\beta}\ln\frac{\sigma_{a+\lambda/2}^{(-\beta)}(x)}{\sigma_{a-\lambda/2}^{(-\beta)}(x)}.
		\end{equation}
		where $x,a,\lambda,\beta\in\mathbb{R},$ $\lambda, \beta \neq 0,$  and $\sigma_d^{(\beta)}(x)$ denotes the logistic function:
		\begin{align}
		\sigma_d^{(\beta)}(x) = \frac1{1+e^{-\beta\cdot (x-d)}}.
		\end{align}
		\label{squashdef}
	\end{thm}
	The Squashing function given in Definition \ref{squashdef} is a continuously differentiable approximation of the generalized cutting function by means of sigmoid functions (see Figure \ref{fig:squash}). By increasing the value of $\beta$, the Squashing function approaches the generalized cutting function. In other words, $\beta$ drives the accuracy of the approximation, while the parameters $a$ and $\lambda$ determine the center and width.
	The error of the approximation can be upper bounded by $\text{constant}/\beta$, which means that by increasing the parameter $\beta$, the error decreases by the same order of magnitude.
	The derivatives of the Squashing function are easy to calculate and can be expressed by sigmoid functions and itself:
	
	\begin{align}
	\frac{\partial S^{(\beta)}_{a,\lambda}(x)}{\partial x} &= \frac1\lambda\left(\sigma_{a-\lambda/2}^{(\beta)}(x)-\sigma_{a+\lambda/2}^{(\beta)}(x)\right) 
	\end{align}
	
	In \cite{why}, it is shown that the formulas describing the squashing functions follow from natural symmetry requirements and contain linear units as a particular case.
	
	\section{Implementation and benchmark tests}\label{S1}
	This section describes the exact implementation of the Squashing function in the PyTorch framework (GitHub Repository \cite{github}).
	To verify this implementation, three experiments were conducted. First, the basics of these experiments are introduced and the datasets are presented.
	The results obtained by using Squashing functions in different benchmark tasks, including tests in different neural network architectures and a comparison with commonly used activation functions, indicate that the Squashing function is capable of performing similarly to %on the same level as 
	other popular activation functions.
	As a starting point, in Section \ref{testing_of_the_squashing_function}, the behavior of the Squashing function with $a=0.5$, $\lambda=1$, and learnable $\beta$ parameter is investigated. 
	\subsection{Testing of the Squashing Function} \label{testing_of_the_squashing_function}
\begin{figure}[!htb]
		\centering
		\includegraphics[width=0.5\textwidth]{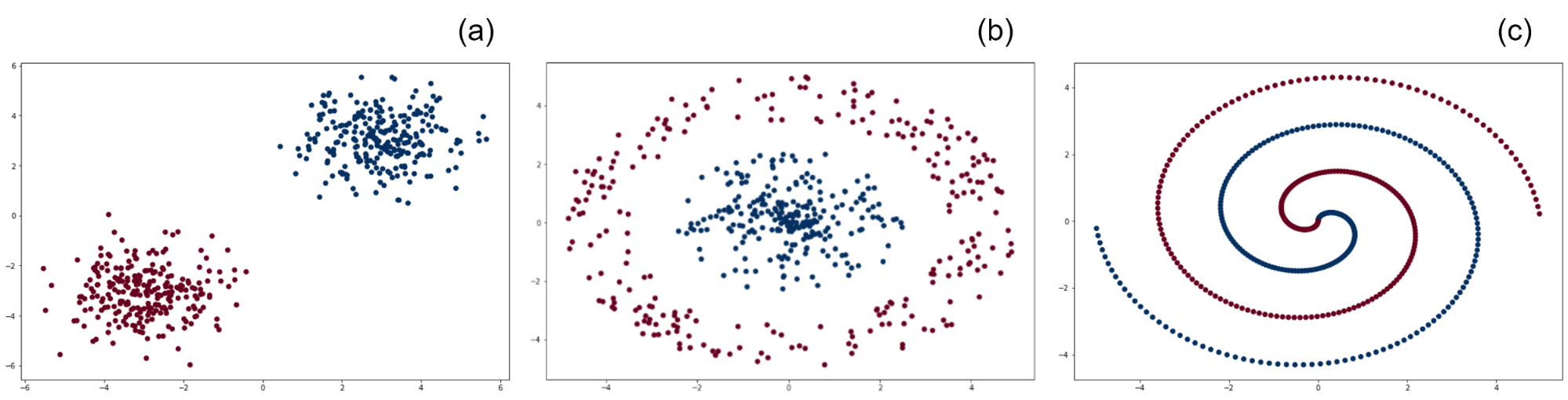}
		\caption{Experimental datasets - (a) Gaussian data, (b) circle data, (c) spiral data}
		%	\source{\footnotesize Own representation} 
		\label{fig:experiment_dataset}
	\end{figure}

	The test phase is divided into three experiments. The goal is to see if the Squashing function could solve simple classification problems.
	Each of these experiments consisted of classifying a set of data that are distributed in different shapes. The dataset is composed of two balanced classes, each containing 250 points. 
	In the first experiment, two point clouds are to be separated by a straight line. In the second experiment, these point clouds are arranged in circular configurations as seen in Figure \ref{fig:experiment_dataset}b. In the last experiment, the point sets formed two intertwined spirals.
	
	\subsubsection{Experiment 1: Classification of Gaussian Data}
\begin{figure}[!htb]
		\centering
		\includegraphics[width=0.5\textwidth]{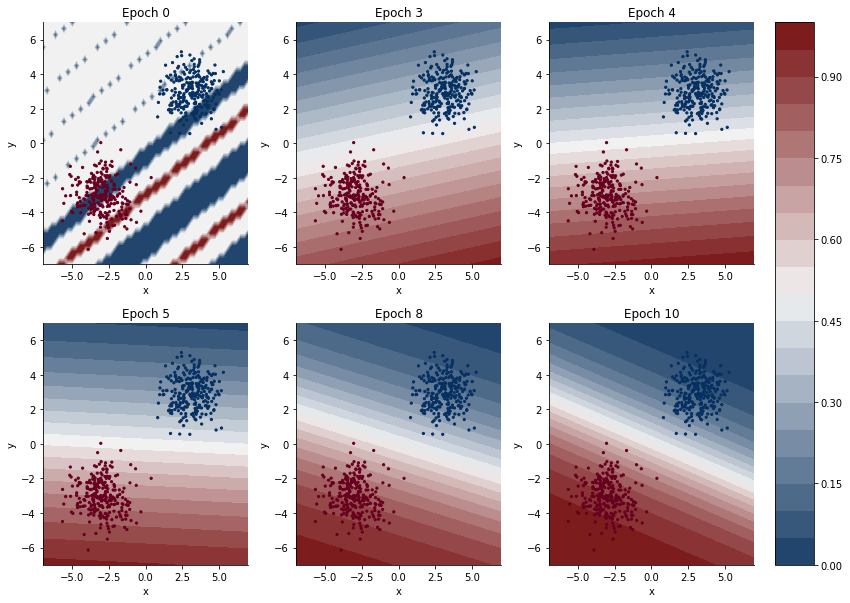}
		\caption{Results of Experiment 1 - Classification of Gaussian data}
		%	\source{\footnotesize Own representation.} 
		\label{fig:experiment1}
	\end{figure}

	The task of the first experiment is solved with a one-layer feedforward network. The model architecture shown in Table \ref{tab:experiment1}  uses a fully-connected layer with two input and two output features.
	\begin{table}
		\centering
		\resizebox{\columnwidth}{!}{%
			\begin{tabular}{|l|l|l|}
				\hline
				\textbf{Type of layer} & \textbf{Number of input features} & \textbf{Number of output features} \\
				\hline
				Fully-connected & 2 & 2 \\
				\hline
			\end{tabular}
		}
		\caption{Experiment 1 - Technical Parameters}
		\label{tab:experiment1}
	\end{table}
	As a cost function cross-entropy function is applied, while the Adam optimization algorithm is utilized for the optimization procedure. The training process takes 10 epochs with a learning rate of $\eta=0.1.$\newline
	Figure \ref{fig:experiment1} shows the visualization of the optimization process for 10 epochs.

	\subsubsection{Experiment 2: Classification of Circle Data}
	
\begin{figure}[!htb]
		\centering
		\includegraphics[width=0.5\textwidth]{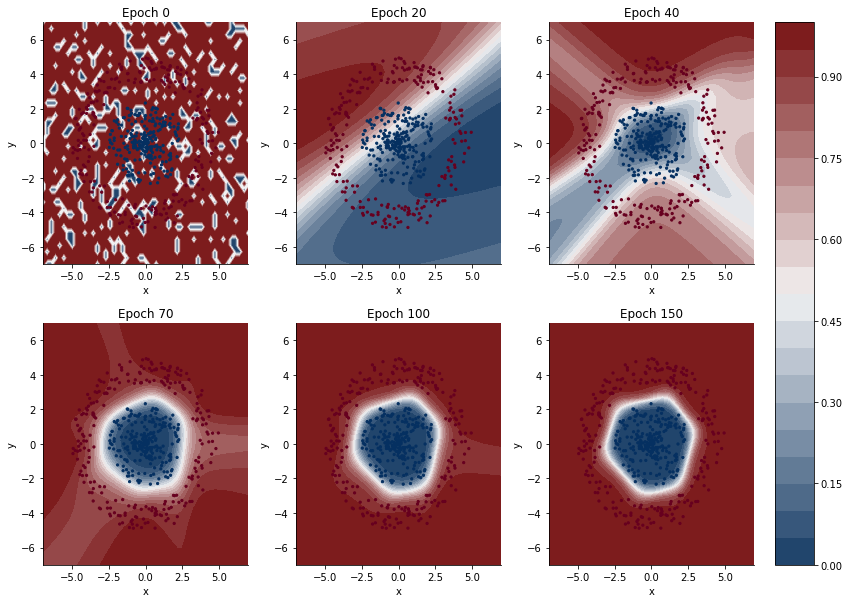}
		\caption{Results of Experiment 2 - Classification of circle data}
		%	\source{\footnotesize Own representation} 
		\label{fig:experiment2}
	\end{figure}

	The problem of the second experiment is solved with a two-layer feedforward network. The model architecture shown in Table \ref{tab:experiment2} uses an input layer with two input features and one output layer with eight input features.
	\begin{table}
		\centering
		\resizebox{\columnwidth}{!}{%
			\begin{tabular}{|l|l|l|}
				\hline
				\textbf{Type of layer} & \textbf{Number of input features} & \textbf{Number of output features} \\
				\hline
				Fully-connected & 2 & 8 \\
				\hline
				Fully-connected & 8 & 2 \\
				\hline
			\end{tabular}
		}
		\caption{Experiment 2 - Technical Parameters}
		\label{tab:experiment2}
	\end{table}
	Similarly to experiment one, a cross-entropy function is applied with the Adam optimization algorithm. The training process takes 150 epochs, with a learning rate of $\eta=0.1$.
	The visualization of the optimization process for 150 epochs can be seen in Figure \ref{fig:experiment2}.
	
	\subsubsection{Experiment 3: Classification of Spiral Data}
\begin{figure}[!htb]
		\centering
		\includegraphics[width=0.5\textwidth]{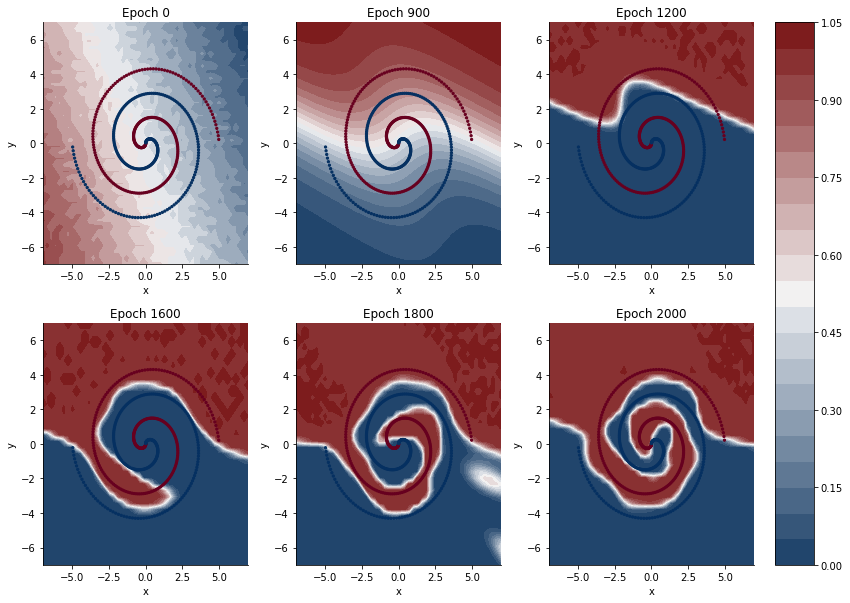}
		\caption{Results of Experiment 3 - Classification of spiral data}
		%	\source{\footnotesize Own representation} 
		\label{fig:experiment3}
	\end{figure}

	In the third experiment, a three-layer feedforward network is employed. The model architecture shown in Table \ref{tab:experiment3} uses an input layer with two input features, one hidden layer with 64 input features, and an output layer with 128 input features. 
	\begin{table}
		\centering
		\resizebox{\columnwidth}{!}{%
			\begin{tabular}{|l|l|l|}
				\hline
				\textbf{Type of layer} & \textbf{Number of input features} & \textbf{Number of output features} \\
				\hline
				Fully-connected & 2 & 64 \\
				\hline
				Fully-connected & 64 & 128 \\
				\hline
				Fully-connected & 128 & 2 \\
				\hline
			\end{tabular}
		}
		\caption{Experiment 3 - Technical Parameters}
		\label{tab:experiment3}
	\end{table}
	Similarly to first two experiments, a cross-entropy function is applied with the Adam optimization algorithm. The training process takes 2000 epochs, with a learning rate of $\eta=0.001$.
	The visualization of the optimization process for 2000 epochs can be seen in Figure \ref{fig:experiment3}.
	\subsubsection{Results} \label{resuts_and_discussion}
\begin{figure}[!htb]
		\centering
		\includegraphics[width=0.5\textwidth]{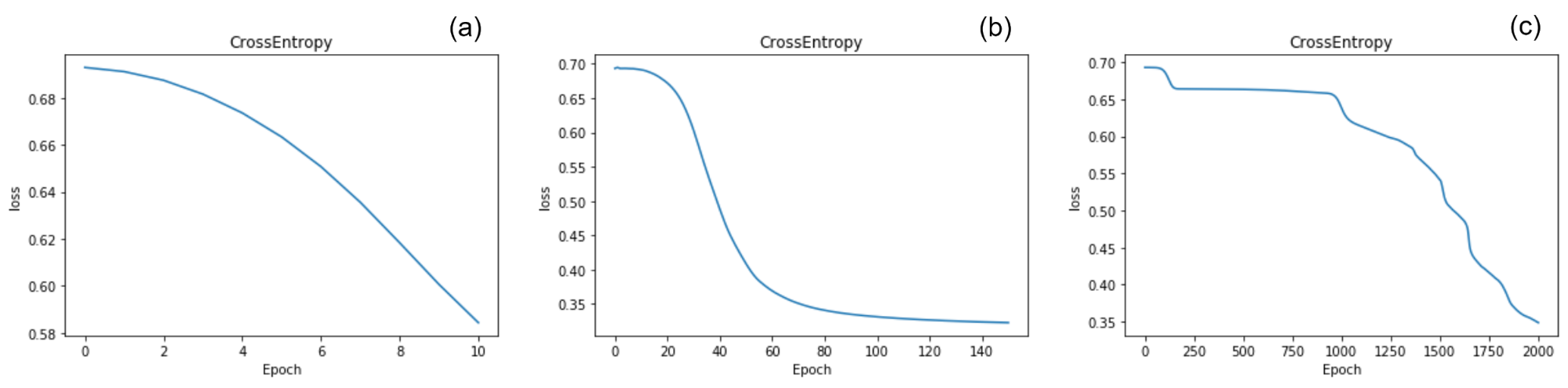}
		\caption{Results Cross-Entropy - (a) Gaussian, (b) circle, (c) spiral}
		%	\source{\footnotesize Own representation.} 
		\label{fig:cross_entropy}
	\end{figure}

	Figure \ref{fig:cross_entropy} shows the learning curves obtained in the experiments, which illustrate the evolution of the cost function for the training set. By observing the loss curves, we can conclude that the Squashing function is capable of solving the tasks of classifying Gaussian, circle, and spiral data. The optimization process of the three experiments clearly shows success in separating both classes. For more computational details see Table \ref{tab:res}.
	
	\begin{table}
		\centering
		\resizebox{\columnwidth}{!}{%
			\begin{tabular}{|c|c|c|c|c|c|}
				\hline
				\textbf{Experiment} & \textbf{$\beta_{\rm{init}}/\beta_{\rm{final}}$} & \textbf{epoch}&\textbf{loss}&\textbf{train accuracy} &\textbf{test accuracy} \\
				\hline
				Gaussian &$0.1/1.4$  &10 &0.58&1& 1\\
				\hline	
				Circle & $10^{-6}/0.565$ & 150 & 0.33 & 1 & 1\\
				\hline	
				Spiral & $0.1/1.4924$ & 2000& 0.36 & 0.96 & 0.94\\
				\hline
			\end{tabular}
		}
		\caption{Determination of $\beta$ parameter in the Gaussian, circle and spiral spatial configurations}
		\label{tab:res}
	\end{table}
	
	\subsection{Benchmarking on FASHION-MNIST}
	In this Section, a benchmark test of various activation functions is presented to compare the performance of the Squashing function with other popular activation functions in a classification problem on FASHION-MNIST, a dataset consisting of 60000 training images and 10000 test images. Each of them is a grayscale image of 28 by 28 pixels in sizes, showing a piece of clothing from Zalando distributed in 10 different categories. In the benchmarks, it should be determined whether the Squashing function can deliver similar performance results as conventional activation functions. 
	The architectures used to solve the classification of the benchmarks tests are: \nameref{lenet}, \nameref{inception_v3}, \nameref{shufflenet_v2}, \nameref{squeezenet}, and \nameref{densenet_121}.
	A more detailed description of the individual networks can be found in Section \ref{networks}.% \nameref{networks}.
	For each network separate runs for the following activation functions was performed: \textit{Rectified Linear Unit (ReLU), Sigmoid function, Hyperbolic tangent (Tanh), Squashing function.}
	Because of the learnable parameter in the Squashing function, the run with this function was performed twice: first with a dynamic, learnable $\beta,$ and then with a static value for $\beta$. Following the same strategy as in the experiments presented in Section \ref{testing_of_the_squashing_function}, %\nameref{testing_of_the_squashing_function}, 
	a cross-entropy function is applied with the Adam optimization algorithm. The value of the learning rate is set to 0.0001 and the size of the batches to 32. The total of amount the training process for each network architecture is 50 epochs. %Note that the number of epochs is capped at 50.%; because free use on the relevant platforms Paperspace and Kaggle did not allow for longer processing times.
	\subsection{Networks} \label{networks}
	\subsubsection{LeNet} \label{lenet}
	The first prototype of the LeNet model was introduced in the year 1989 by Yann LeCun et al  \cite{Le98b}. They combined a Convolutional Neural Network trained by backpropagation algorithm to learn the convolution kernel coefficients directly from images.
	This prototype was able to recognize handwritten ZIP code numbers for the United States Postal Service and became the foundation of Convolutional Neural Networks.
	A few years later, in 1998,  LeCun et al. published a paper about gradient-based learning applied to document recognition, in which they reviewed different methods of recognizing handwritten characters on paper and used standard handwritten digits to identify benchmark tasks \cite{Le98a}. The results showed that the network exceeded all other models.
	The most common form of the LeNet-Model is the LeNet-5 Architecture. The LeNet-5 is a seven-layer neural network architecture (excluding inputs) that consists of two alternate convolutional and pooling layers followed by three fully connected layers (dense layers) at the end \cite{Zh20}. This network was successfully used in ATM check readers which could automatically read the check amount by recognizing hand-written numbers on checks. 
	
	%\begin{figure}
	%\centering
	%\includegraphics[width=.75\textwidth]{images/06_benchmarks/lenet.png}
	%\caption{Visualization of the LeNet-5 architecture}
	%	\source{\footnotesize \cite{Zh20}}
	%\label{fig:dataset_production}
	%\end{figure}
	
	\subsubsection{Inception-v3} \label{inception_v3}
	The Inception-v3 network was proposed by a research group at Google in 2015 and is a 42-layer deep learning network with higher computational efficiency and fewer parameters included compared to other state-of-the-art CNN networks \cite{Sz16}. 
	%The third version was a development that was released at the same time as the se\-cond version and uses some other training algorithms \cite{Sz16}. 
	With about 24 million parameters, this network is one of the largest and most computationally intensive during the benchmarks.
	Inception-v3 uses so-called Inception Modules. These act as multiple filters that are applied to the same input value by means of convolution layers and pooling layers. By using different filter sizes, different patterns can be extracted from the input images that increases the number of trainable parameters. This procedure increases memory consumption and computing time considerably, however leads to a significant increase in accuracy. %\cite{Sz16} Figure \ref{fig:inception_v3} shows a schematic structure of the network architecture. 
	%\begin{figure}
	%\centering
	%\includegraphics[width=0.75\textwidth]{images/06_benchmarks/inception-v3_overview.png}
	%\caption{Visualization of the Inception-v3 architecture}
	%\source{\footnotesize \url{https://cloud.google.com/tpu/docs/inception-v3-advanced}}
	%\label{fig:inception_v3}
	%\end{figure}
	\subsubsection{ShuffleNet-v2} \label{shufflenet_v2}
	ShuffleNet, published in 2018 by Ma et al. \cite{Ma18}, also seeks to improve efficiency, but is designed for mobile devices with limited computing capabilities. The improvement in efficiency is given by the introduction of two new operations: point-wise group convolution and channel shuffle. 
	%\begin{figure}
	%\centering
	%\includegraphics[width=0.8\textwidth]{images/06_benchmarks/shufflenet_v2.png}
	%\caption{Visualization of the ShuffleNet-v2 architecture}
	%	\source{\footnotesize \cite{Ma18}}
	%\label{fig:shufflenet-v2}
	%\end{figure}
	The main drawback of 1x1 convolutions, also known as point-wise convolutions, is the relative high computational cost that can be reduced by using group convolutions. The channel shuffle operation has shown to be able to mitigate some unintended side effects that may evolve.
	In general, the group-wise convolution divides the input feature maps into two or more groups in the channel dimension and performs convolution separately on each group. It is the same as slicing the input into several feature maps of smaller depth and then running a different convolution on each.
	After the grouped convolution, the channel shuffle operation rearranges the output feature map along the channels dimension.
	
	%\begin{figure}
	%	\centering
	%\includegraphics[width=.60\textwidth]{images/06_benchmarks/shufflenet_shuffle.png}
	%\caption{Visualization of the shuffle operation}
	%	\source{\footnotesize \cite{Ho20}}
	%\label{fig:shuffle_operation}
	%\end{figure}
	
	\subsubsection{SqueezeNet} \label{squeezenet}
	SqueezeNet, which was developed in 2016 within the cooperation of DeepScale, University of California, University of Berkeley, and Stanford University, is a convolutional neural network architecture proposed by Iandola et al. \cite{La16} that seeks to achieve levels of accuracy similar to previous architectures, while significantly reducing the number of parameters in the model. SqueezeNet relies primarily on reducing the size of the filters by combining channels to decrease the inputs of each layer and to handle larger feature maps. This yields to better feature extraction despite the reduction in the number of parameters. This optimization of the feature extraction is done by applying subsampling to these maps at the final network layers, rather than after each layer. 
	The basic building block of SqueezeNet is called the Fire module. It is composed of a squeeze layer that is in charge of input compression consisting of 1x1 filters. These combine all channels of each input pixel into one. It has also an expand layer which combines 3x3 and 1x1 filters for feature extraction. %\cite{La16} The architecture of SqueezeNet is shown in Figure \ref{fig:squeezenet}.
	
	%\begin{figure}
	%\centering
	%\includegraphics[scale= 0.7]{images/06_benchmarks/squeezenet.png}
	%\caption{Visualization of the SqueezeNet architecture}
	%	\source{\footnotesize \cite{La16}}
	%\label{fig:squeezenet}
	%\end{figure}

	\subsubsection{DenseNet-121} \label{densenet_121}
	The main goal of the DenseNet-121, which was released in 2015 by Facebook AI Research, is to reduce the model size and complexity \cite{Ch17}.
	In Dense convolution networks, each layer of the feature map is concatenated with the input of each successive layer within a dense block. This allows later layers within the network to directly leverage the features from earlier layers, encouraging feature reuse within the network \cite{Jj18}. Concatenating feature maps learned by different layers increases the variation in input from subsequent layers, improving efficiency. As the network is able to use any previous feature map directly, the number of parameters required can be reduced considerably \cite{Hu16}.
	
	%\begin{figure}
	%\centering
	%\includegraphics[width=.40\textwidth]{images/06_benchmarks/densenet-121.png}
	%\caption{Visualization of a 5-layer dense block}
	%	\source{\footnotesize \cite{Hu16}}
	%	\label{fig:densenet-121}
	%\end{figure}
	\subsection{Results and Discussion} \label{benchmark_results}
	In this section, the results of the benchmarking on FASHION-MNIST is demonstrated. For each network listed in Section \ref{networks}, separate runs for the following activation functions is performed: ReLU, Sigmoid function, Tanh, Squashing function with a static (squashing-nl, $\beta_{\rm{initial}}=0.1$), as well as with a dynamic, learnable $\beta$ parameter. 
	\subsubsection{LeNet}
\begin{figure}[!htb]
		\centering
		\includegraphics[width=0.5\textwidth]{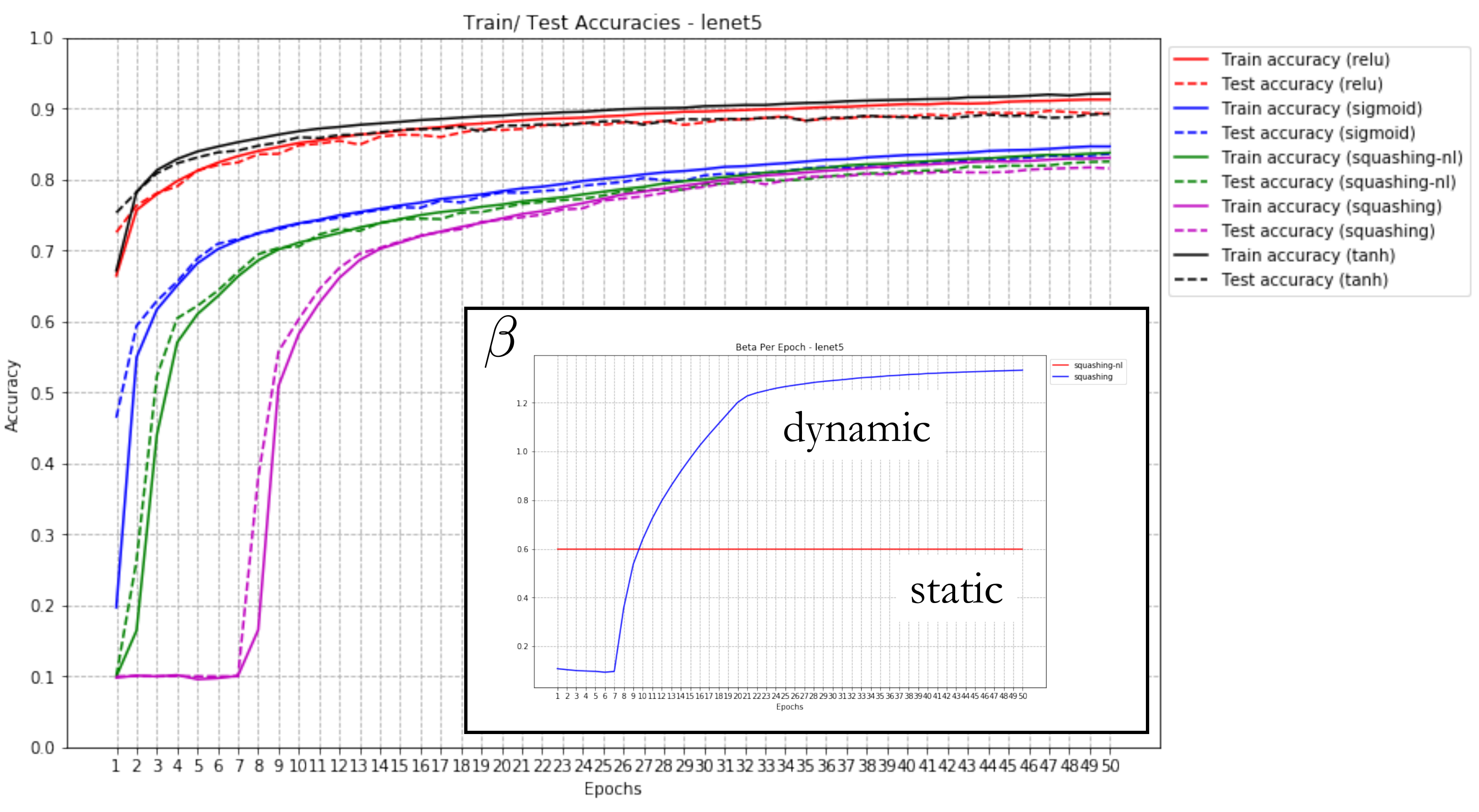}
		\caption{Line plot showing learning curves of accuracies for different activation functions applied in the Lenet-5 architecture}
		%	\source{\footnotesize Own representation.}
		\label{fig:lenet5_acc}
	\end{figure}

	The accuracy over a period of 50 epochs is shown in Figure \ref{fig:lenet5_acc}. The Squashing function with an adjustable $\beta$ parameter has an accuracy of 10 \% until epoch 7, then rises steeply and settles at 81\%. Note that the training of the Squashing function with a learnable beta parameter needs more initial steps to approximate the appropriate $\beta$ parameter value. The inset of Figure \ref{fig:lenet5_acc}  displays the course and adjustment of the $\beta$ value for the Squashing function with dynamic and static $\beta$ values. Despite the larger computational cost, this additional procedure strengthens the veracity of the applied method.  %For the squashing function with dynamic $\beta$, the value starts to increase sharply at epoch 7 and is about 1.30 at epoch 50. %The value of the static beta stays at 0.6 over the 50 epochs.
	The accuracy curves of both Squashing functions (with dynamic $\beta$ and static $\beta$) and of the sigmoid function settle at about 81\%. In contrast, the accuracy of the activation functions ReLU and Tanh reaches 91\%. The trends for the test and training process are similar. 
	
\begin{figure}[!htb]
		\centering
		\includegraphics[width=0.5\textwidth]{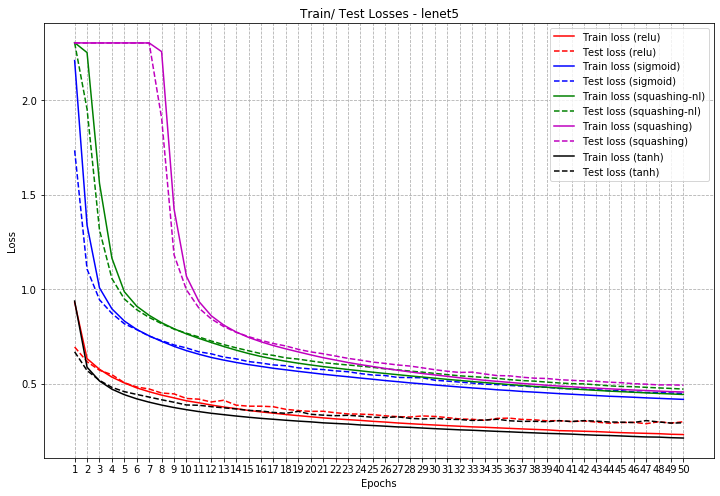}
		\caption{Line plot showing learning curves of loss for different activation functions applied in the Lenet-5 architecture}
		%	\source{\footnotesize Own representation.}
		\label{fig:lenet5_loss}
	\end{figure}

	Figure \ref{fig:lenet5_loss} illustrates the course of the loss value for the different activation functions. The value of the loss converges towards 0. The deviation between the training and the test loss is negligible for all activation functions. Consequently, the network has the ability to make predictions even for unseen datasets. No overfitting or underfitting takes place here.
	
\begin{figure}[!htb]
		\centering
		\includegraphics[width=0.5\textwidth]{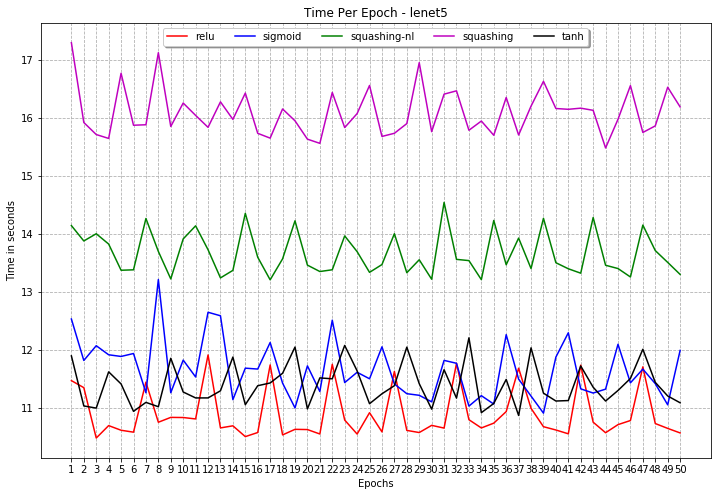}
		\caption{Line plot showing the time performance for different activation functions applied in the Lenet-5 architecture}
		%	\source{\footnotesize Own representation.}
		\label{fig:lenet5_time}
	\end{figure}

	Figure \ref{fig:lenet5_time} demonstrates the runtime in seconds for the different activation functions, as a function of the number of epochs.
	It is noticeable that the Squashing function with adjustable $\beta$ value takes between 15.5 and 17 seconds per epoch. In comparison, the other activation functions (squashing-nl included) perform somewhat better. However, this difference can be compensated by the fact that the Squashing function has the potential of modeling nilpotent logic.
	
	%\begin{figure}
	%\centering
	%\includegraphics[width=0.5\textwidth]{images/06_benchmarks/lenet-5_beta.png}
	%\caption{Adjustment of the beta value per epoch applied in the Lenet-5 architecture}
	%	\label{fig:lenet5_beta}
	%\end{figure}
	
	\subsubsection{Inception-v3}
\begin{figure}[!htb]
		\centering
		\includegraphics[width=0.5\textwidth]{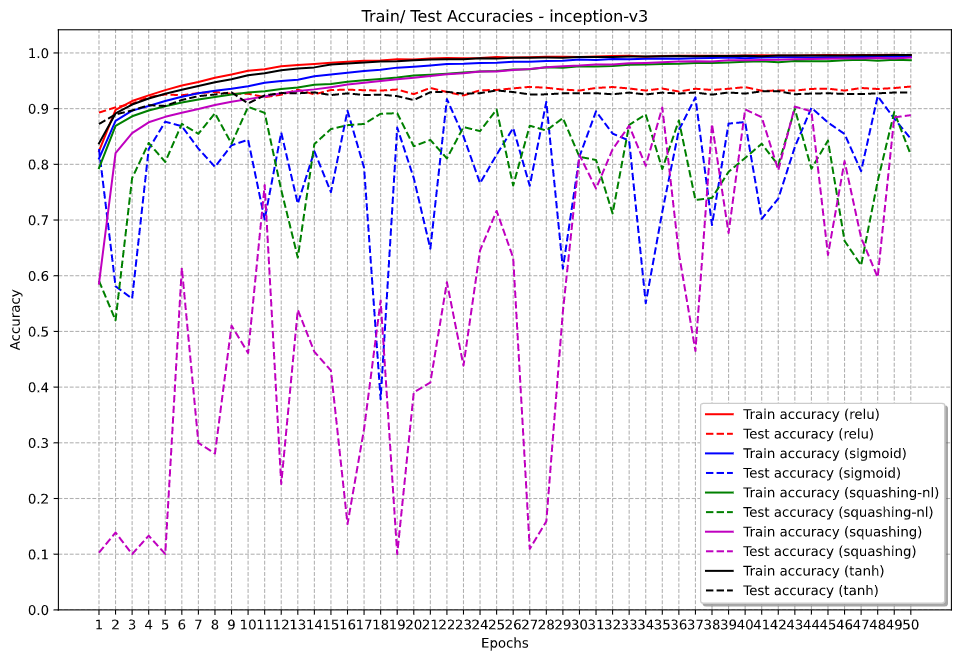}
		\caption{Learning curves of accuracies for different activation functions applied in the Inception-v3 architecture}
		%	\source{\footnotesize Own representation.}
		\label{fig:inception-v3_acc}
	\end{figure}

\begin{figure}[!htb]
		\centering
		\includegraphics[width=0.5\textwidth]{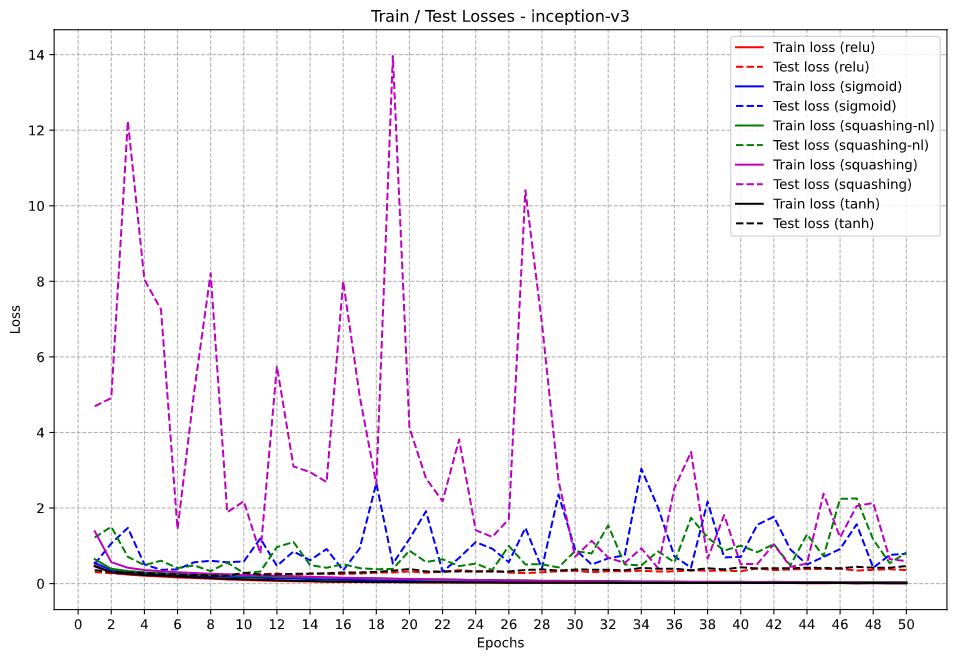}
		\caption{Learning curves of loss for different activation functions applied in the Inception-v3 architecture}
		%	\source{\footnotesize Own representation.}
		\label{fig:inception-v3_loss}
	\end{figure}

	Similar to Figure \ref{fig:lenet5_acc}, Figure \ref{fig:inception-v3_acc} provides information about the accuracy of the investigated activation functions over a time period of 50 epochs for the network Inception-v3. %The x-axis illustrates the number of training steps in epochs; the y-axis the accuracy of the training and test set.
	A special characteristic that stands out is the significant fluctuation of the test accuracy in case of the Squashing, the Squashing-nl and the sigmoid function. This indicates difficulties in making predictions, although the train accuracy for all activation functions is above 90\%. However, the amplitude of this waving effect decreases after a couple of tens of epochs landing at above 85\% at epoch 50.  Note here that with about 24 million parameters, this network is one of the largest and most computationally intensive during the benchmarking. This can explain the initially fluctuating behavior. As a consequence, the development of the loss behaves similarly as shown in Figure \ref{fig:inception-v3_loss}. Note the performance of Squashing-nl being close to that of the other activation functions. %What is interesting, however, is that the test accuracies of the three activation functions mentioned above settle at a high accuracy value of about 88\% at epoch 50.
	%The diagram does not show why these big jumps happen. This is a typical case of a black-box model as it is very difficult to interpret such a problem when they occur. However, there are so-called confusion matrices, %(see \nameref{confusion_matrix}), 
	%which are used to better understand the prediction.
	
	%Figure \ref{fig:inception-v3_loss} shows the development of the loss over 50 epochs. %The extreme fluctuation of the test loss from the squashing, the squashing-nl and the sigmoid functions in comparison to the ReLU and the Tanh function are clearly recognizable. %Normally, the course of the loss should be continuous and flat as can be observed in the train loss of all activation functions. It would be very interesting to analyze how the loss values of the different activation functions behave after 50 epochs. This could not be tested in this work due to the limitations of the runtimes for training a network on the platforms Kaggle, Google Colab and Paperspace.
	
	The graphs  "time per epoch" and "Beta per epoch" can be found in the Appendix.
	
	\subsubsection{ShuffleNet-v2}
\begin{figure}[!htb]
		\centering
		\includegraphics[width=0.5\textwidth]{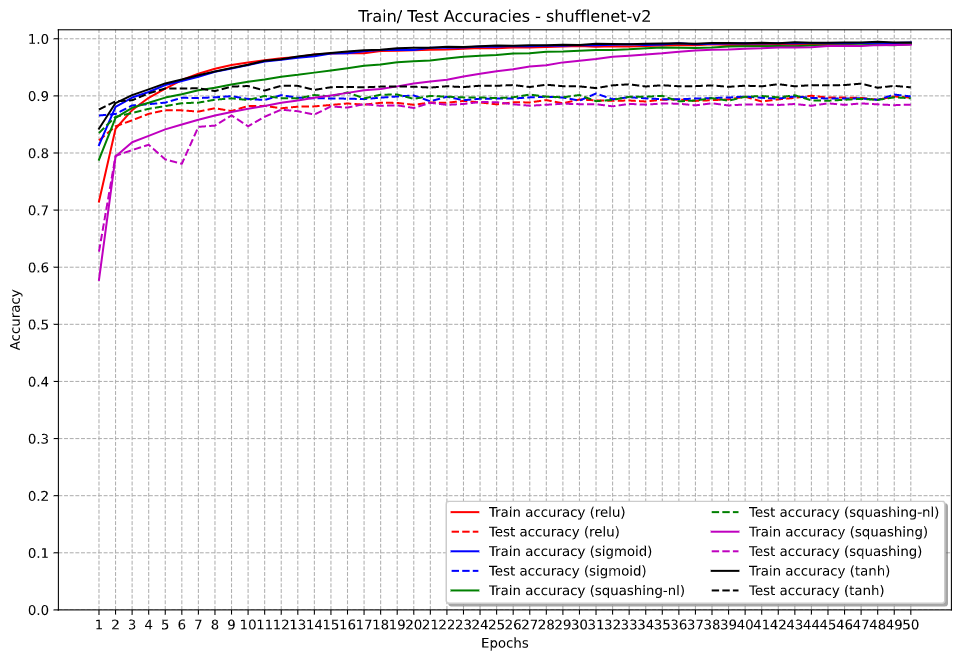}
		\caption{Learning curves of accuracies for different activation functions applied in the ShuffleNet-v2 architecture}
		%	\source{\footnotesize Own representation.}
		\label{fig:shufflenet-v2_acc}
	\end{figure}

	Similar to Figure \ref{fig:lenet5_acc} and \ref{fig:inception-v3_acc}, Figure \ref{fig:shufflenet-v2_acc} provides information about the accuracy of the investigated activation functions over a time period of 50 epochs for the network ShuffleNet-v2. The accuracy for the train and the test set of the different activation functions shows a steady development. %(Figure \ref{fig:shufflenet-v2_acc}). 
	Compared to ReLU, Sigmoid and Tanh functions, the train accuracy of the Squashing and Squashing-nl function increases slower but settles above 98\% accuracy like the other functions. Surprisingly, the different activation functions show also very high test accuracy values of about 90\% at epoch 50.
	
\begin{figure}[!htb]
		\centering
		\includegraphics[width=0.5\textwidth]{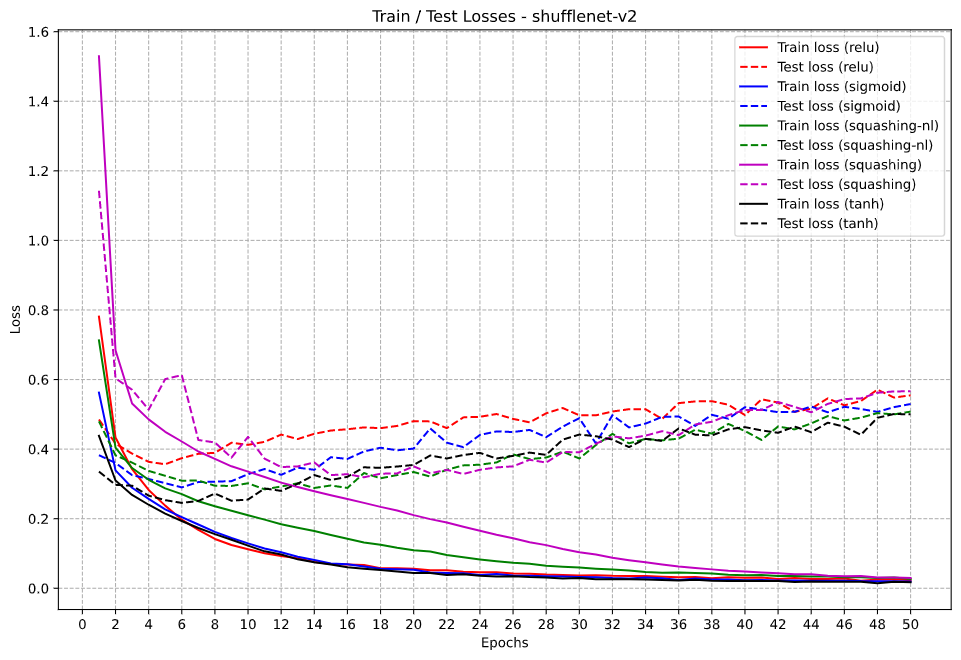}
		\caption{Learning curves of loss for different activation functions applied in the ShuffleNet-v2 architecture}
		%	\source{\footnotesize Own representation.}
		\label{fig:shufflenet-v2_loss}
	\end{figure}

	In Figure \ref{fig:shufflenet-v2_loss}), with respect to the loss, the network overfits for each activation function. This is indicated by the fact that test and the train loss show an increasing deviation.
	
	The graphs  "time per epoch" and "Beta per epoch" can be found in the Appendix.
	
	\subsubsection{SqueezeNet}
\begin{figure}
		\centering
		\includegraphics[width=0.5\textwidth]{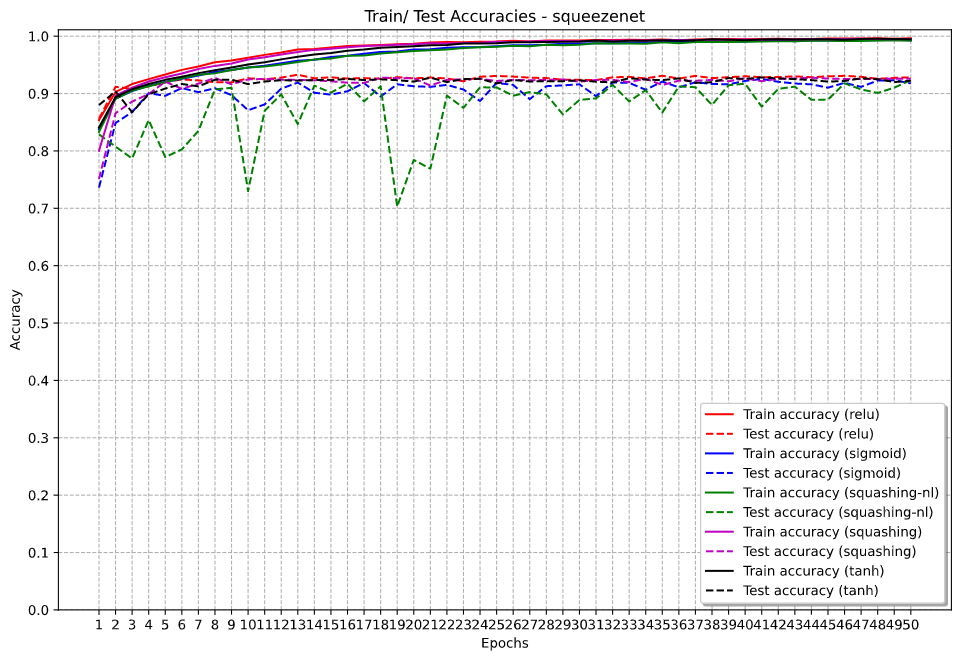}
		\caption{Learning curves of accuracies for different activation functions applied in the SqueezeNet architecture}
		%	\source{\footnotesize Own representation.}
		\label{fig:squeezenet_acc}
	\end{figure}

	The SquezzeNet accuracy diagram given in Figure \ref{fig:squeezenet_acc} illustrates that the progression of the training and test set curves is similar to that of ShuffleNet-v2. %Noticeable are the jumps of the dotted line for the squashing-nl function. This is because the network cannot accurately predict the test set data.
	
	The diagram for the losses given in Figure \ref{fig:squeezenet_acc} clearly illustrates %that the difficulties not only lie in the large jumps of the squashing-nl but also in the increasing deviation of the dotted lines (test loss) and the continuous lines (train loss) demonstrating 
	that the network is overfitting for all of the examined activation functions. 
	
	\begin{figure}
		\centering
		\includegraphics[width=0.5\textwidth]{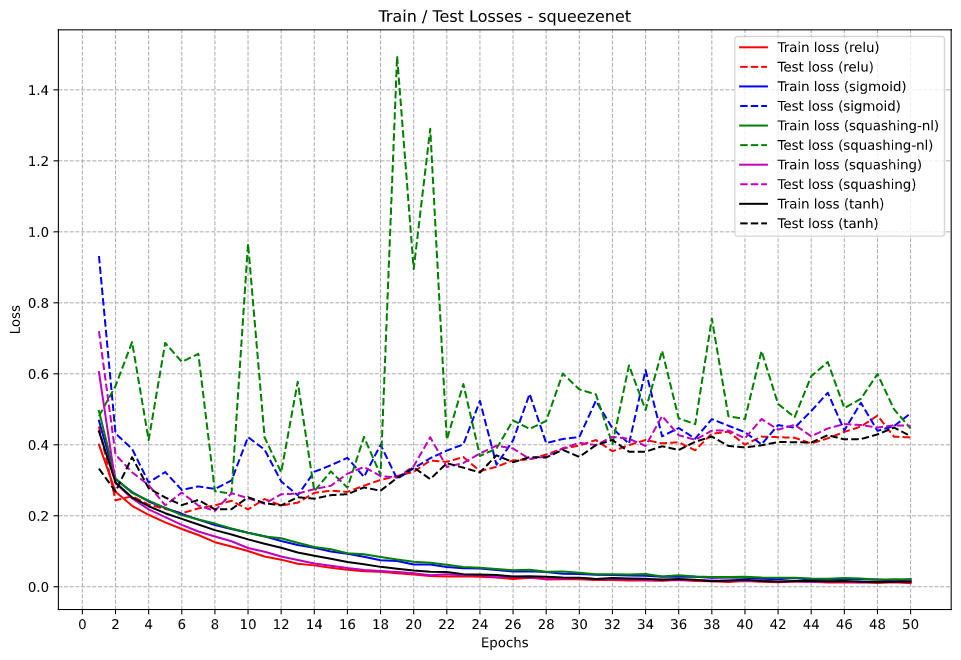}
		\caption{Learning curves of loss for different activation functions applied in the SqueezeNet architecture}
		%	\source{\footnotesize Own representation.}
		\label{fig:squeezenet_loss}
	\end{figure}
	
	The graphs  "time per epoch" and "Beta per epoch" can be found in the Appendix.
	
	\subsubsection{DenseNet-121}
\begin{figure}
		\centering
		\includegraphics[width=0.5\textwidth]{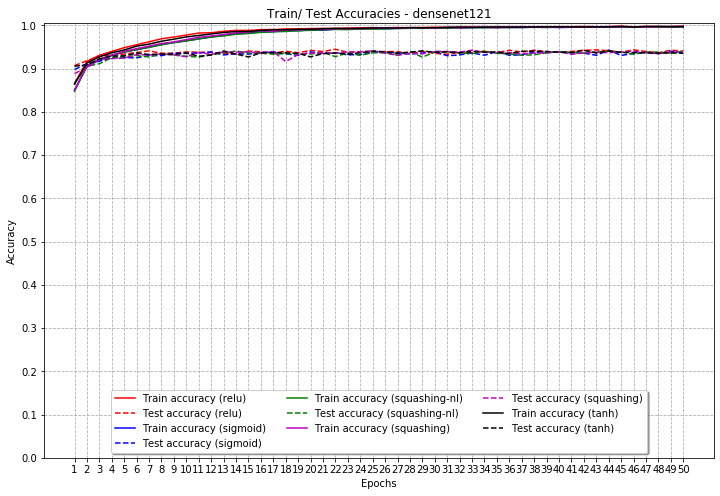}
		\caption{Learning curves of accuracies for different activation functions applied in the DenseNet-121 architecture}
		%	\source{\footnotesize Own representation.}
		\label{fig:densenet-121_acc}
	\end{figure}

	The accuracy diagram of the Densenet-121 plotted in Figure \ref{fig:densenet-121_acc} demonstrates the development of the accuracy over 50 epochs. It is notable that there is no difference in the train accuracies of the different activation functions. The same characteristics applies to the test set for all the activation functions used in this network. The train accuracy for all activation functions lies at about 99\% and the test accuracy at about 94\%.
	
\begin{figure}
		\centering
		\includegraphics[width=0.5\textwidth]{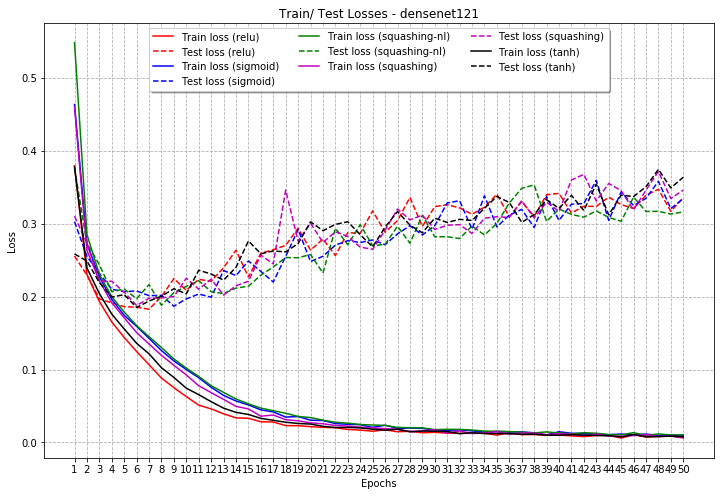}
		\caption{Learning curves of loss for different activation functions applied in the DenseNet-121 architecture}
		%	\source{\footnotesize Own representation.}
		\label{fig:densenet-121_loss}
	\end{figure}

	In the losses diagram of the Densenet-121 in Figure \ref{fig:densenet-121_loss}, the large deviation between test and train loss is particularly visible. This deviation causes the network to overfit. %Therefore, it can be defined that if new datasets which the network has never seen before are used for predictions, they will be predicted wrong.
	
	The graphs  "time per epoch" and "Beta per epoch" can be found in the Appendix.

	\subsubsection{Evaluation in terms of confusion matrices} \label{confusion_matrix}
	A confusion matrix is a tool that allows one to see the performance of a model in a general way, where each column of the matrix represents the identification class that the model predicts, while each row represents the expected class, the true input. The diagonal indicates which images were correctly predicted.
	One of the advantages of a confusion matrix is that they make it easier to see which categories the network is confusing with one another. It is usually used in supervised learning.
	The prediction accuracy and classification error can be calculated as follows \cite{Vi11}:
	\begin{equation} \label{eq_accuracy}
	Accuracy = \frac{total\  correct\  predictions}{total\  predictions\  made} \cdot 100
	\end{equation}
	
	\begin{equation} \label{eq_error}
	Error = \frac{total\  incorrect\  predictions}{total\  predictions\  made} \cdot 100
	\end{equation}
	
\begin{figure}[!htb]
		\centering
		\includegraphics[width=0.45\textwidth]{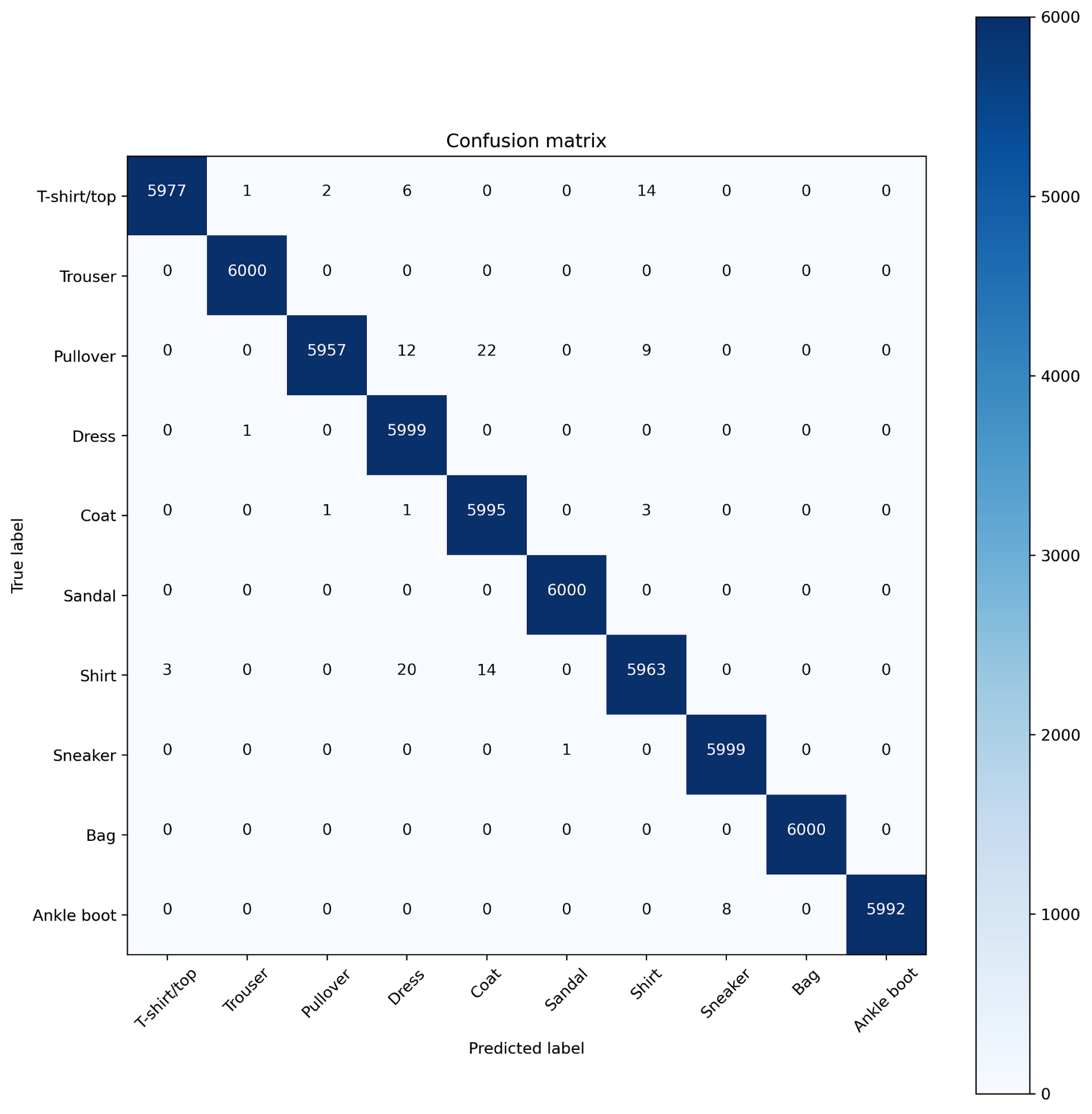}
		\caption{Confusion matrix for the training set of the DenseNet-121}
		%	\source{\footnotesize Own representation.}
		\label{fig:cm_train}
	\end{figure}

\begin{figure}[!htb]
		\centering
		\includegraphics[width=0.45\textwidth]{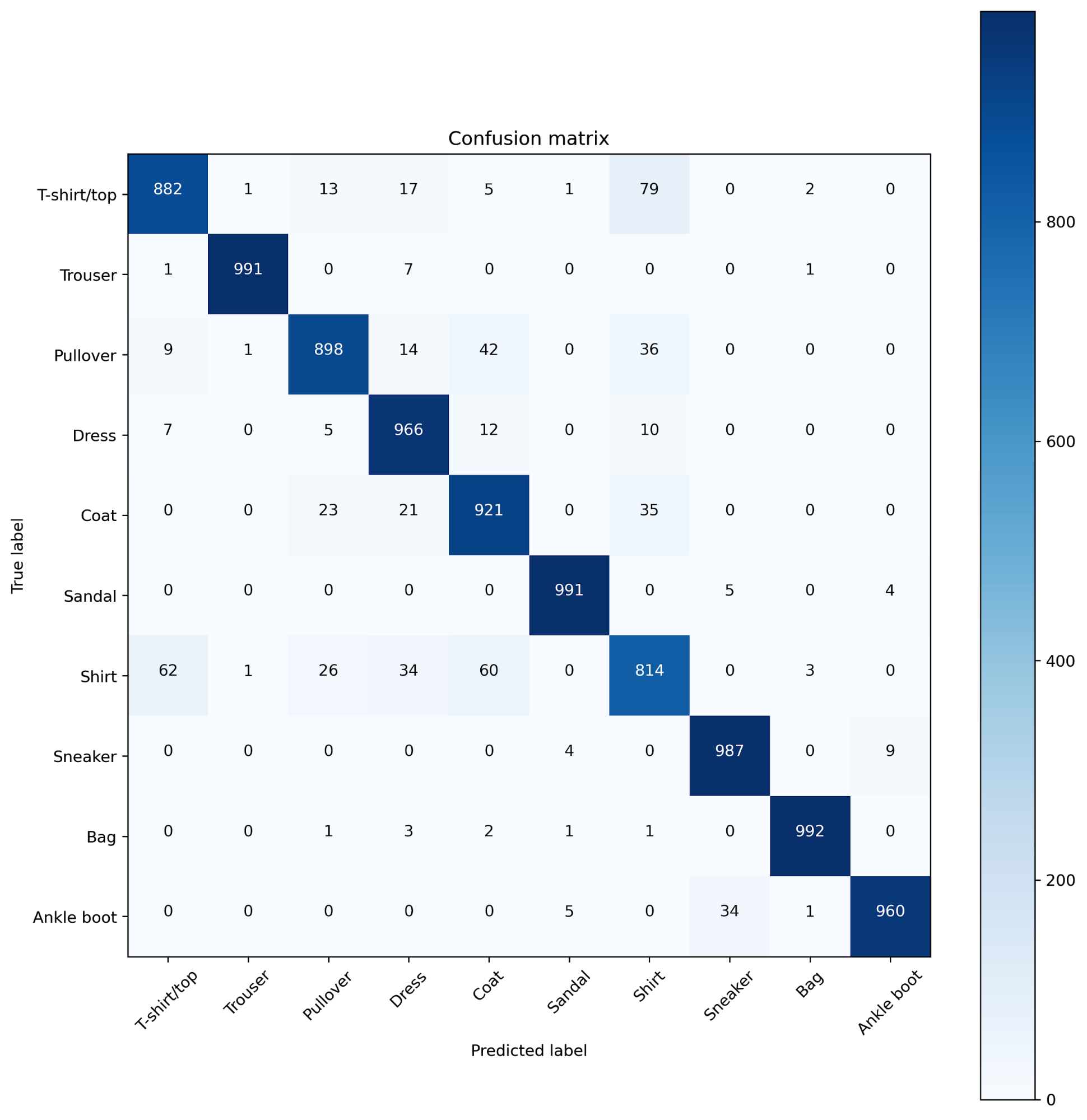}
		\caption{Confusion matrix for the test set of the DenseNet-121}
		%	\source{\footnotesize Own representation.}
		\label{fig:cm_test}
	\end{figure}

	Figures \ref{fig:cm_train} and \ref{fig:cm_test} display an example of a confusion matrix for the train and test sets of the DenseNet-121. The distribution of the total dataset consists of 10 classes. The training accuracy of the Squashing function in the Densenet-121 is 99.8\%, while the test accuracy is 94\%.

	%By applying the formula \ref{eq_accuracy}, we get:
	\begin{equation} \label{eq_accuracy_densenet-121-1}
	Train\ Accuracy = \frac{59882}{60000} \cdot 100=99.803\ \%
	\end{equation}
	\begin{equation} \label{eq_accuracy_densenet-121}
	Test\ Accuracy = \frac{9402}{10000} \cdot 100=94.02\ \%
	\end{equation}
	
	%Figure \ref{fig:cm_test} shows an example of a confusion matrix for the test set of the DenseNet-121.
	
	%To confirm this the formula \ref{eq_accuracy} was again applied resulting in the following:

	The confusion matrices for each network and the corresponding activation functions can be found in the Appendix.
	
	\section{Implementing nilpotent logical gates in neural networks}\label{S2}
	As we have seen, nilpotent logical systems provide a suitable mathematical background for the combination of continuous nilpotent logic and neural networks, contributing to the improvement of interpretability and safety of machine learning. The following sections describe the implementation of the conjunction, one of the most important continuous logical operators. According to Table \ref{operators}, the conjunction can be modeled by $\left[x+y-1\right]$; i.e. by a perceptron with fixed weights ($w_i=1$),  fixed bias ($C=-1$) and the cutting function or its differentiable approximation, the Squashing function as activation function.
	%Discrete logic values can take only two states: 0 or 1 (False or True). In continuous logic, on the other hand, values can take continuous values in a compact interval such as [0, 1]. \cite{Ya10} This is the output in neural networks if the input values are normalized between 0 and 1. These values can be transformed into their logical values by an activation function, e.g. the cutting function \cite{Do05}, within the nilpotent logical operators.
	
	%To translate values to an interval a continuous function c: $[0,1]^n \to [0,1]$ is needed. The cutting function itself is not continuous and therefore not suitable for this use case. However, it is possible to approximate the cutting function with the squashing function with a high value for beta e.g. greater than 50. This can be seen in Figure \ref{fig:squashing_function_50}.
	
	%%To employ continuous logical operators, two neurons are first connected with a logical AND gate. For two-dimensional input values, each neuron has two weights. These weights form the gradient $m$ and the y-intercept $n$ of a straight line with the equation $y = mx + n$. These straight lines can be connected in many ways and thus classified as complex point sets. %The implementation of this procedure is described in this section.
	
\begin{figure*}[!t]
		\centering
		\includegraphics[width=0.7\textwidth]{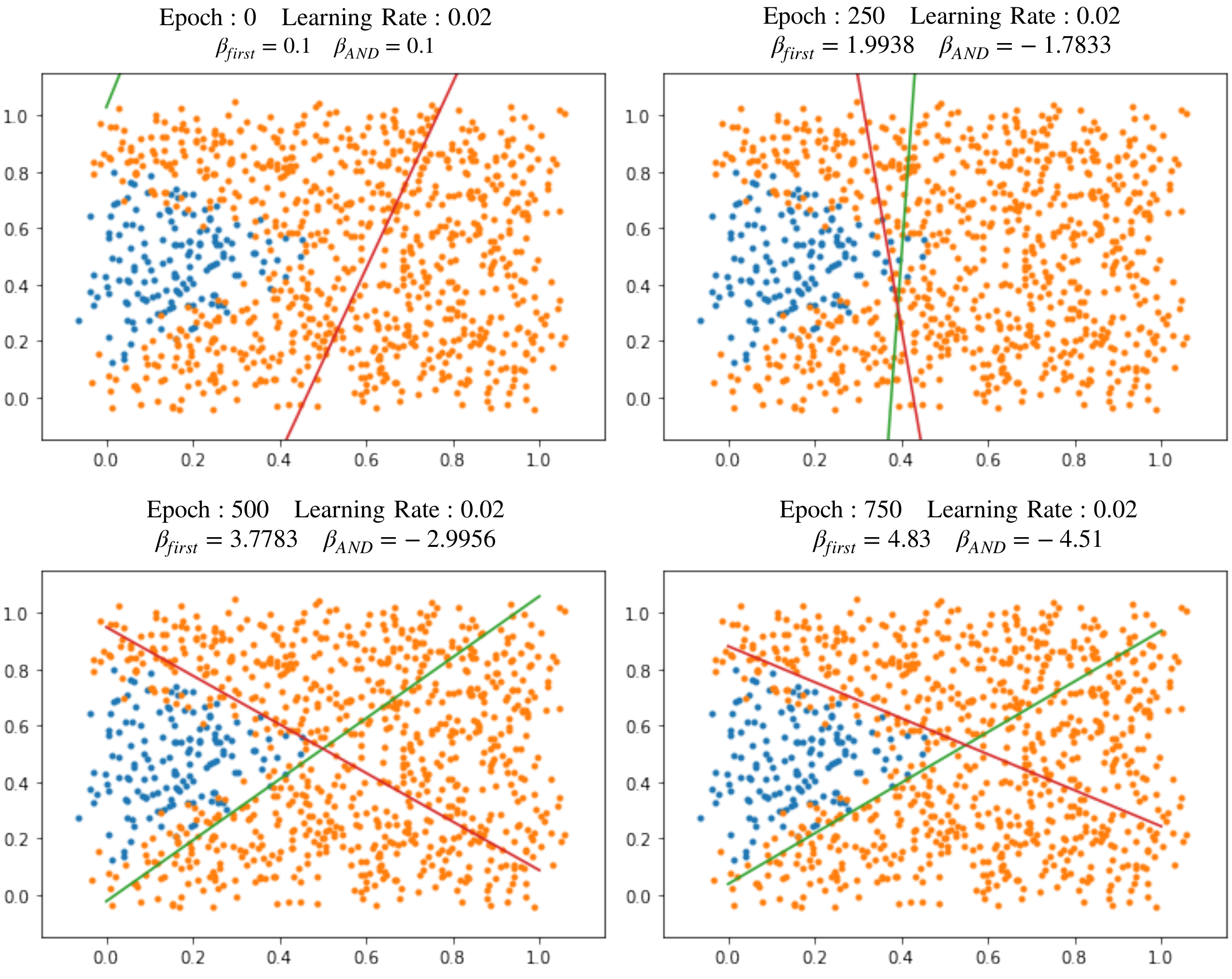}
		\caption{Results of 750 training epochs with a two-line shallow network}
		\label{fig:shallow_network_2_results}
	\end{figure*}

	As a first experiment shown in Figure \ref{fig:shallow_network_2_results}, we define a classification problem where two intersecting straight lines delineate a segment of the plane to be found by a shallow network. This segment is defined by
	\begin{equation} b_1\,y\geq m_1\, x+c_1 \quad \text{AND} \quad b_2\,y\leq m_2\, x+c_2.\label{and}\end{equation}
	
	Here, not only the AND operator, but also the inequalities can be modeled by perceptrons. The output values should be the truth values of the inequalities. A perceptron with weights $-m_1, b_1,$ and bias $-c_1$ in case of the first inequality, and $m_2, -b_2,$ and bias $c_2$ in case of the second one, using the cutting (or its approximation, the Squashing) activation function can model the soft inequalities well:
	
	\begin{equation} [b_1\,y-m_1\, x-c_1] \quad \text{AND} \quad [m_2\, x-b_2\,y+c_2].\label{soft}\end{equation}
	
\begin{figure}[!htb]
		\centering
		\includegraphics[width=0.4\textwidth]{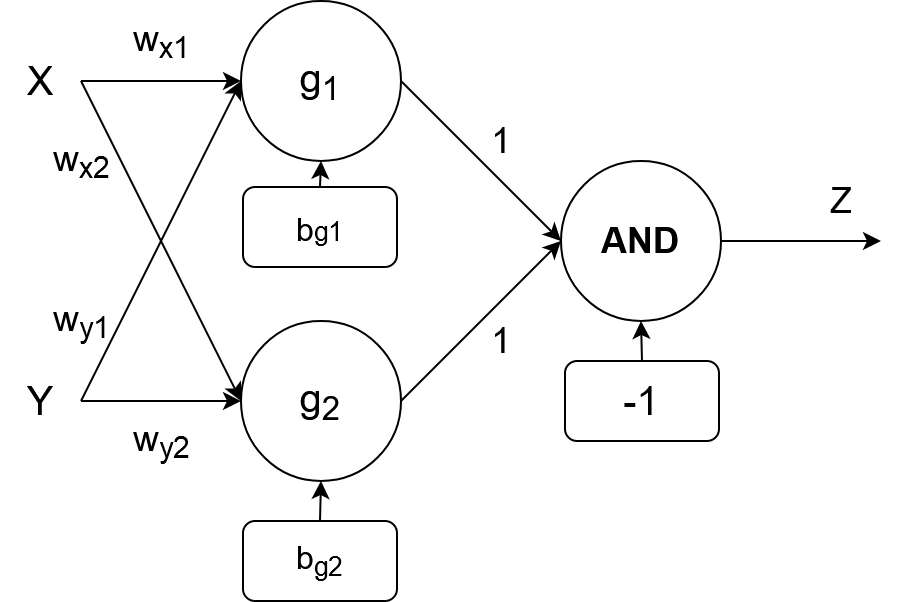}
		\caption{Schematic representation of a shallow network with two neurons}
		%	\source{\footnotesize Own representation} 
		\label{fig:shallow_network_2}
	\end{figure}

	This means, to model the problem described in Equation \eqref{soft}, a shallow network with only two layers needs to be set up (see Figure \ref{fig:shallow_network_2}). The weights and biases in the first layer are to be learned, while the parameters of the hidden layer are frozen (modeling the conjunction). This architecture is similar to that of Extreme Learning Machines (ELM) introduced by Huang et al. in \cite{elm}, where the parameters of hidden nodes need not be tuned. ELMs are able to produce good generalization performance and learn thousands of times faster than networks trained using backpropagation. The model suggested here can combine extreme learning machines with the continuous logical background, being a promising direction towards a more interpretable, transparent, and safe machine learning.
	
	%an attempt was made to divide a point set with two straight lines and classify the points that way. For this, a shallow network with only two neurons was set up. In this network, the two neurons each represent straight lines whose gradient $m$ is represented by the weights $w_{xi}$ or $w_{yi}$ and whose y-axis section is represented by the corresponding weights $w_{xj}$ or $w_{yj}$.
	
\begin{figure}[!htb]
		\centering
		\includegraphics[width=0.5\textwidth]{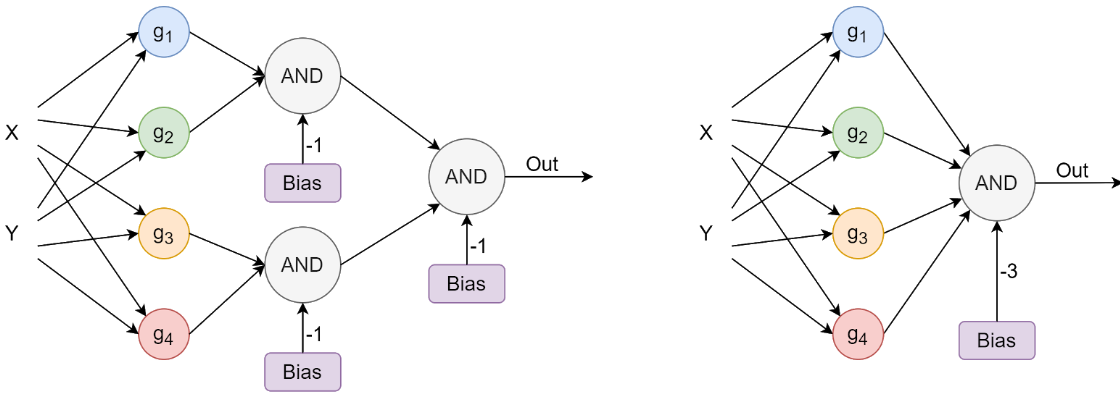}
		\caption{AND gates: connecting four neurons}
		%\source{\footnotesize Own representation} 
		\label{fig:four_connection}
	\end{figure}

	After this implementation, the number of straight lines is increased to four. If more straight lines are connected by AND gates, more AND gates with two inputs are required. However, these can be reduced to one gate after the bias has been adapted to it. This connection is shown in Figure \ref{fig:four_connection}.
	
	\begin{equation}
	\text{Out} = \overbrace{[\underbrace{[g_1 + g_2 - 1]}_{AND} + \underbrace{[g_3 + g_4 - 1]}_{AND} - 1]}^{AND} = \underbrace{[g_1 + g_2 + g_3 + g_4 - 3]}_{AND}
	\end{equation}
	
	%The implementation inside the code is rather simple. Inside the \lstinline{__init__}method two new linear layers have to be implemented and the bias in line 14 has to be changed to the corresponding number of linear layers minus one. These layers have to be introduced inside the forward function as well. Considering these simple changes, this procedure can later be implemented dynamically in the network and can be adapted automatically as the number of neurons increases.
	
	\subsection{Experiment 1: Two Lines}
	For the first experiment, a dataset is divided into two categories. An open angular shape is labeled with 1 (blue) at the edge of the data field. The remaining data points are labeled 0 (orange). The goal is to separate the two datasets with two straight lines. The network architecture is predesigned according to the nilpotent model described in Section \ref{S2}. The activation function is the Squashing function with learnable $\beta$ parameter, different for the AND gate in the hidden layer and for the first layer. The learning rate is set to 0.02. %With the AND operation, two classes can be distinguished. 
	After 750 epochs, with a runtime of a few seconds, the network is able to separate the two datasets. A longer runtime further reduces the error. The results can be found in Figure \ref{fig:shallow_network_2_results}. %With each new run, the straight lines are redefined. This means that a line does not always represent the same neuron. 
	
	It is important to note that the processing time of these networks is extremely fast due to their low complexity.
	
	\subsection{Experiment 2: Four Lines}
\begin{figure*}[!t]
		\centering
		\includegraphics[width=0.7\textwidth]{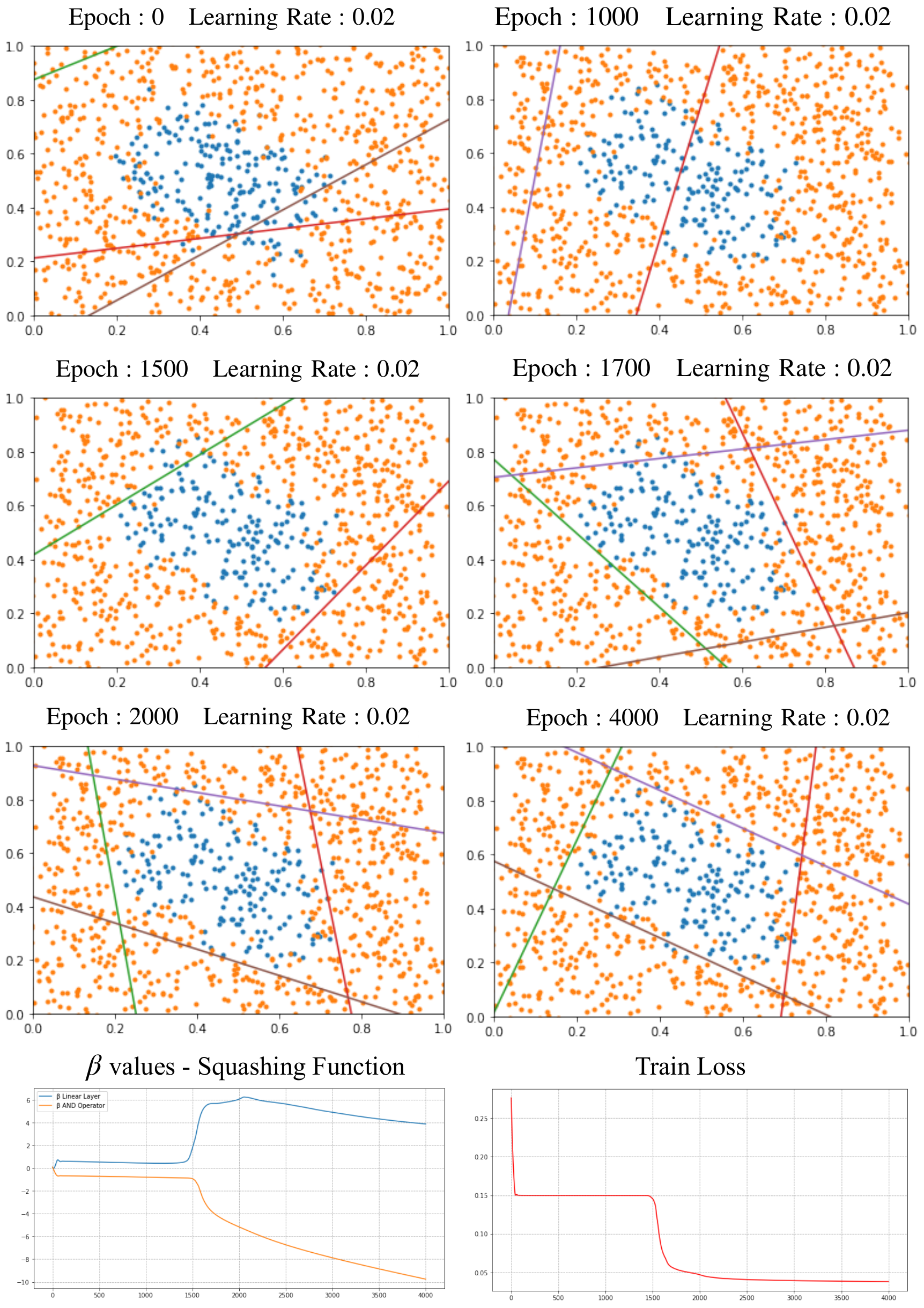}
		\caption{Results of 4000 training epochs with a four-line shallow network with $\beta$ and loss values of 4000 training epochs with a four line shallow network}
		\label{fig:shallow_network_4_results}
	\end{figure*}

	In the second experiment, the generated dataset is similar to the first. A trapezoidal area lies in the middle of the dataset and the data points are labeled with 1 (blue) and 0 (orange). This area can now be separated by four straight lines.
	With about 4000 epochs, the training lasts disproportionately longer than the training with two neurons, although the number of parameters only slightly more than doubled. %This may be due to the higher complexity of the form of the data area. %However, the calculation time of this network remains in the range of significantly less than one minute. 
	The activation function is the Squashing function with learnable $\beta$ parameter. We allow $\beta$ in the AND gate (hidden layer) to be different from that in the first layer. The learning rate is set to 0.02.
	After about 1700 epochs, the network is able to align the four straight lines to the record. Between 2000 and 4000 epochs, the network improves accuracy significantly, adjusting parameters of the straight lines to obtain a more accurate classification (see Figure \ref{fig:shallow_network_4_results}).
	During the development of the $\beta$ parameters of the Squashing functions, the values for the first and for the second layer develop in different directions. Note that allowing $\beta$ to be negative leads to a decreasing activation function (see Figure \ref{fig:squash}). For the interpretation of the hidden layer as a logical gate, a negative $\beta$ value means that in Equation \eqref{soft}, the cutting function is replaced by its decreasing counterpart (a step function with value 1 for negative inputs and value 0 for non-negative ones), which corresponds to finding the complement of the intersection. Clearly, for a binary classifier, finding the intersection is equivalent to finding its complement.
	%Furthermore, the two $\beta$ parameters develop erratically from epoch 1500 onwards. 
	The development of the $\beta$ parameters is illustrated in Figure \ref{fig:shallow_network_4_results}.
	%\begin{figure}
	%\centering
	%\includegraphics[width=0.5\textwidth]{images/07_nilpotent_neural_networks/beta_nilpotent_neural_network.png}
	%\caption{$\beta$ values of 4000 training epochs with a four line shallow network}
	%\label{fig:betas_nilpotent_network}
	%\end{figure}
	%The development of the $\beta$parameter goes along with that of the error. This means that w
	With the corresponding development of the $\beta$ parameter, the error in the network decreases. %In the experiment, there is a jump at epoch 1500 and the error decreases noticeably. A corresponding minimum was found. 
	The development of network loss is displayed in Figure \ref{fig:shallow_network_4_results}.
	
	%\begin{figure}
	%\centering
	%\includegraphics[width=0.5\textwidth]{images/07_nilpotent_neural_networks/train_loss.png}
	%\caption{Loss values of 4000 training epochs with a four-line shallow network}
	%\label{fig:loss_nilpotent_network}
	%\end{figure}
	
	\subsubsection{Other Activation Functions}
\begin{figure}[!htb]
		\centering
		\includegraphics[width=.50\textwidth]{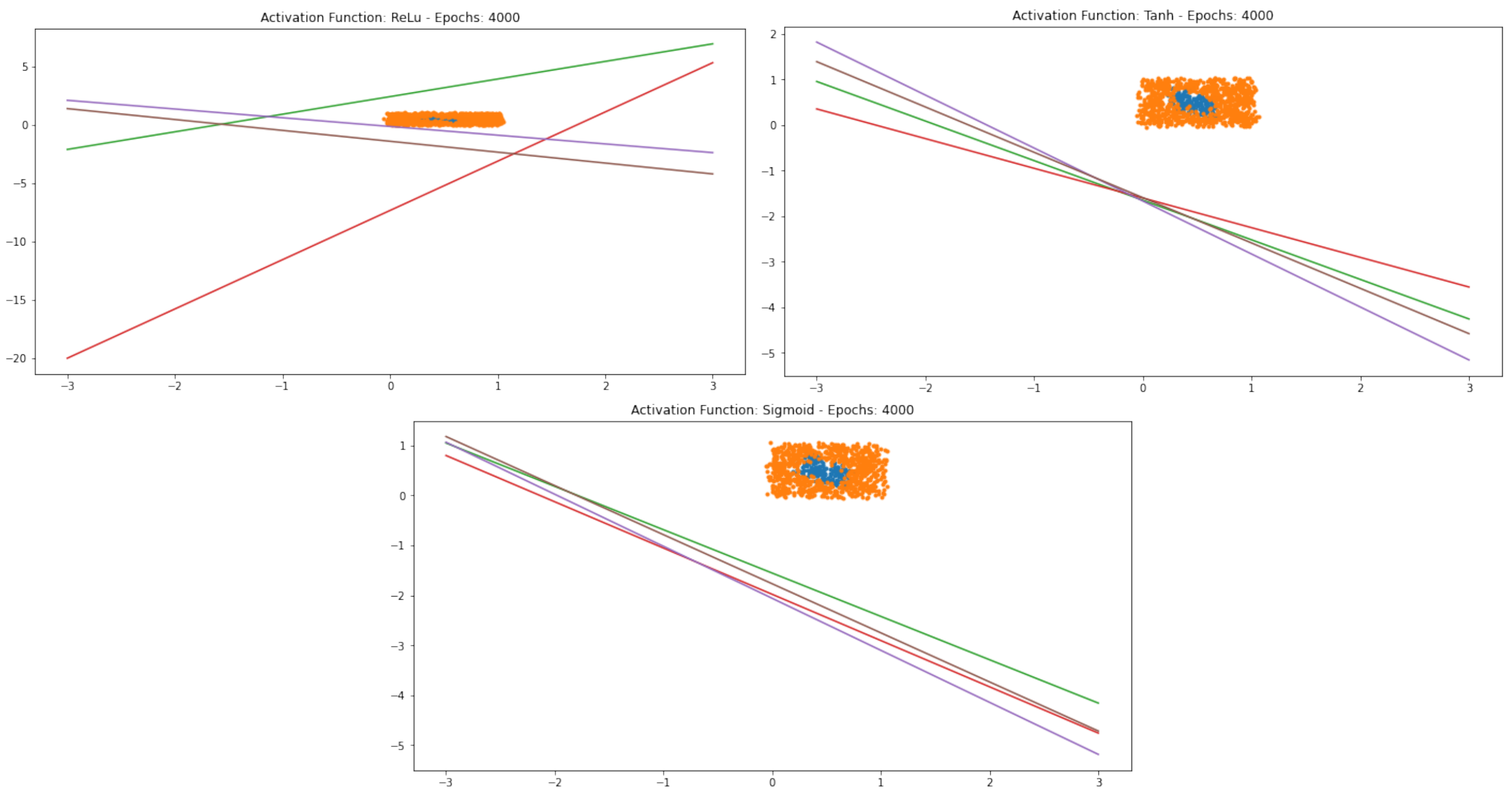}
		%activation_relu.png}
		%\includegraphics[width=.45\textwidth]{images/07_nilpotent_neural_networks/activation_sigmoid.png}\\
		%\includegraphics[width=.45\textwidth]{images/07_nilpotent_neural_networks/activation_tanh.png}
		\caption{Performance of other common activation functions finding a rectangular area using the nilpotent neural model}
		\label{fig:different_activation_functions}
	\end{figure}

	Looking at the other usual activation functions for this application, it stands out that no sufficient results could be achieved in this experiment. %There can be several reasons for this. ReLu and Leaky ReLU are not differentiable, furthermore, they have no upper limit for values above 1 and are therefore not sufficient for modeling continuous logic. \cite{Ca20}.
	%\hl{Sigmoid and tangent hyperbolicus are differentiable, but still need a shift in x-direction and also in y-direction. This is a limitation because the standard implementation of both functions of the framework has to be changed.}
	For the behavior during training with ReLU, sigmoid, and TanH, see Figure \ref{fig:different_activation_functions}. 

\begin{figure}[!htb]
		\centering
		\includegraphics[width=.45\textwidth]{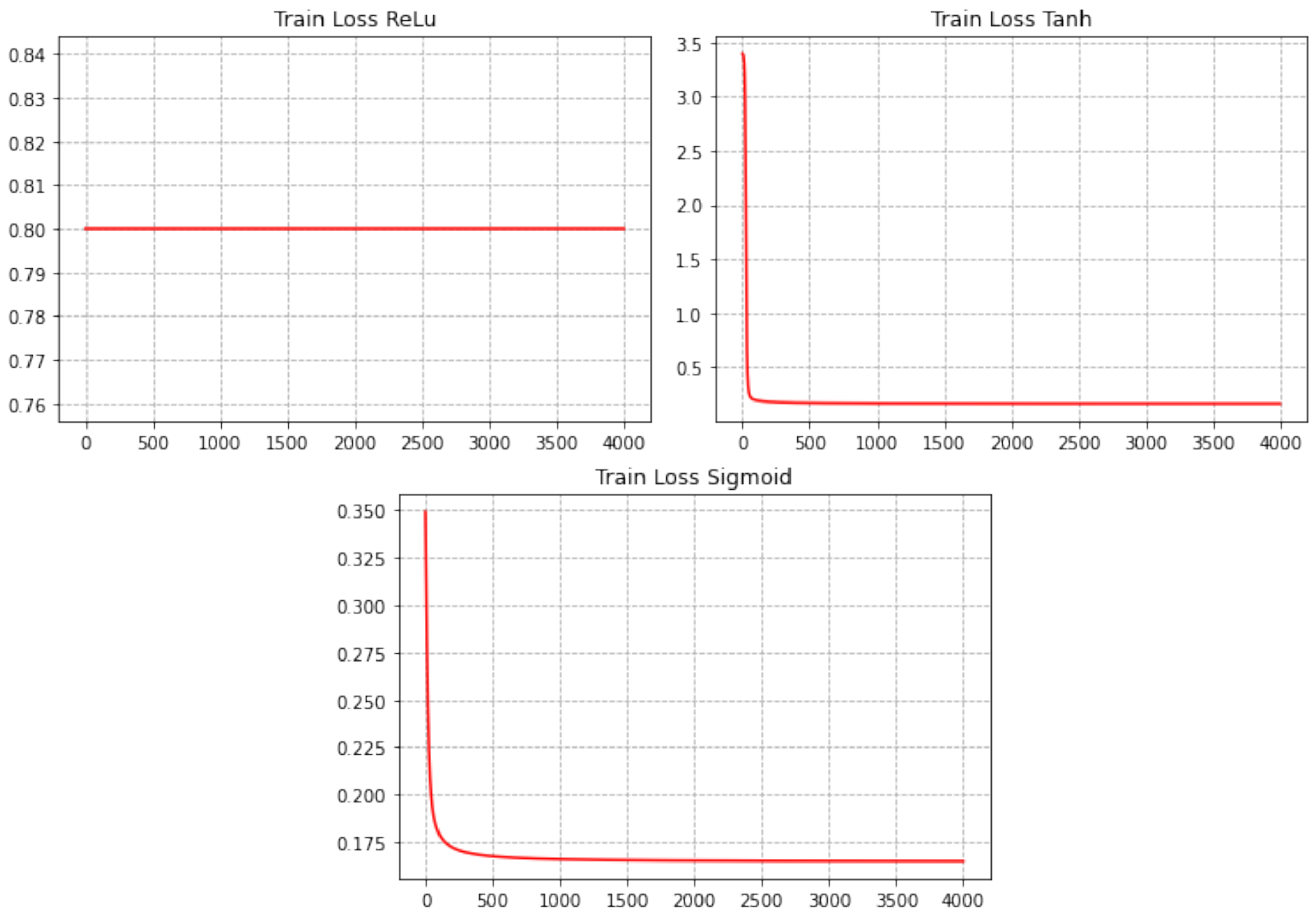}
		%loss_relu.png}
		%\includegraphics[width=.45\textwidth]{images/07_nilpotent_neural_networks/loss_sigmoid.png}\\
		%\includegraphics[width=.45\textwidth]{images/07_nilpotent_neural_networks/loss_tanh.png}
		\caption{Loss of the different activation function}
		\label{fig:loss_activation_functions}
	\end{figure}

	Considering the loss of the individual activation functions, the ReLU function does not improve accuracy. The error remains constant during the entire training period. Using sigmoid or TanH improve in accuracy and the error initially decreases, but this value settles down after a few epochs and then remains almost constant. This development is reflected in Figure \ref{fig:loss_activation_functions}.
	
	\section{Conclusion}\label{Concl}%There can be several reasons for this. ReLu and Leaky ReLU are not differentiable, furthermore, they have no upper limit for values above 1 and are therefore not sufficient for modeling continuous logic.
	As recent research shows, the idea of achieving eXplainable Artificial Intelligence (XAI) by combining neural networks with continuous logic is a promising way to approach the problem of interpretability of machine learning: by this combination, the black-box nature of neural models can be reduced, and the neural network-based models can become more interpretable, transparent, and safe. This hybrid approach suggests using Squashing functions (continuously differentiable approximations of cutting functions) as activation functions. To the best of our knowledge, there has been no attempt in the literature to test the performance of these functions so far. The goal of this study was to implement Squashing functions in neural networks and to test them by conducting benchmark tests. Additionally, we also conducted the first experiments implementing continuous logical gates using the Squashing function.
	
	The implementation of the squashing function was successfully performed with the framework PyTorch and tested with a series of selected experiments and benchmark tests. %The experiments testing continuous logical gates were based on the tool \textit{Tensorflow playground}.
	The aim of the benchmark tests was:
	\begin{enumerate}
		\item to compare the Squashing function with other activation functions,
		\item to test the performance of the activation functions under different conditions, i.e. to measure the performance for different architectures of neural networks. 
	\end{enumerate}
	
	The benchmark tests showed that the performance of the Squashing function is comparable to conventional activation functions. The following activation functions were considered: the Rectified Linear Unit (ReLu), the sigmoid function, the hyperbolic tangent (TanH), and the Squashing function, both with static and with learnable $\beta$ parameter.
	The measured values were determined for the following network architectures: LeNet-5, Inception-v3, ShuffleNet-v2, SqueezeNet and DenseNet-121.
	
	%We can conclude that the Suashing function performed well in the benchmark tests. 
	%Problems occurred during the implementation of the squashing function in PyTorch. It was difficult to find resources for the implementation of a new activation function in PyTorch as there are very few guides or documentation available online.
	%Furthermore, there were also problems when training the networks with the squashing function.  Due to unknown reasons, the $\beta$ value of some training operations took on the value "NaN" and gave unusable results. In addition, there were difficulties with the runtimes for the training processes of the above-mentioned network architectures. Due to the training time limitations of the platforms Kaggle, Google Colab and Paperspaces, the training epochs had to be limited to 50 iterations so that the training process would not be automatically interrupted.
	
	%It is known from different findings that neural networks is an emerging computer technology that can be used in a large number and variety of applications. They can be developed in a reasonable amount of time and perform concrete tasks better than other conventional technologies. Despite that humans find it difficult to fully trust a machine.
	Another focus of this study was the implementation of continuous logic using the Squashing function. The experiments have proven that by utilizing the differentiability of the Squashing function, there is a possible way to implement continuous logic into neural networks, as a crucial step towards more transparent machine learning. 
	
	As a next step, we are working on a comparison with extreme learning machines (ELM) introduced in \cite{elm}, where, similarly to the model suggested in this study, the parameters of hidden nodes are frozen, and need not be tuned. ELMs are able to produce good generalization performance and learn thousands of times faster than networks trained using backpropagation. Combining extreme learning machines with the continuous logical background can be a very promising direction towards a more interpretable, transparent, and safe machine learning. Supplemental research is also in progress aiming to investe which “And”- and “Or”-operations can be represented by the fastest (i.e., 1-Layer) neural networks, and which activations functions allow such representations \cite{XAI}.

	\bibliography{references}
	\bibliographystyle{elsarticle-num}
\section*{Appendix}
\begin{landscape}
	\begin{figure}
		\centering
		\includegraphics[width=25cm]{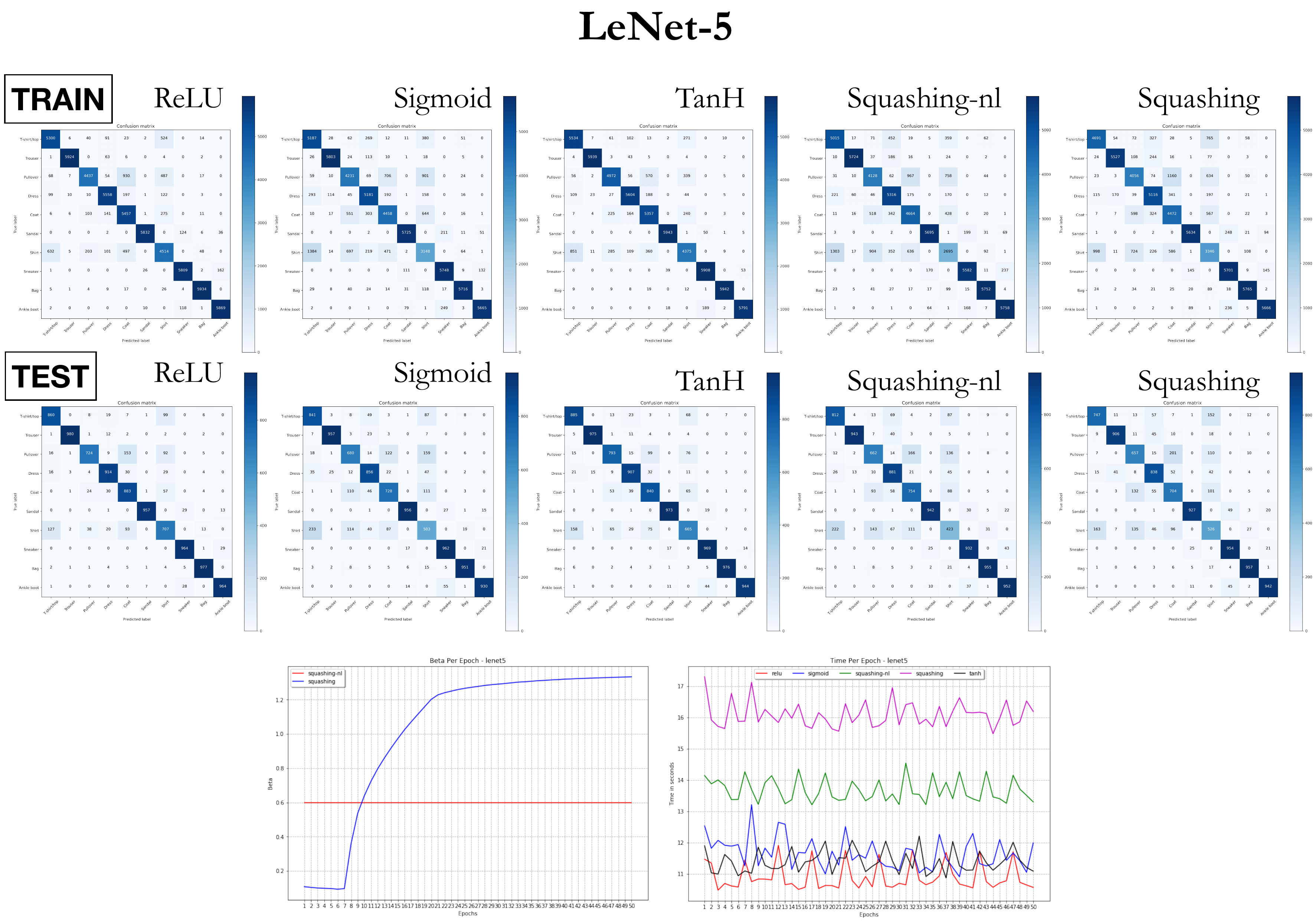}
		%	\caption{Line plot showing the time performance for different activation functions applied in the LeNet-5 architecture}
		%   \source{\footnotesize Own representation.}
		%	\label{fig:inception-v3_time}
	\end{figure}
\end{landscape}

\begin{landscape}
	\begin{figure}
		\centering
		\includegraphics[width=25cm]{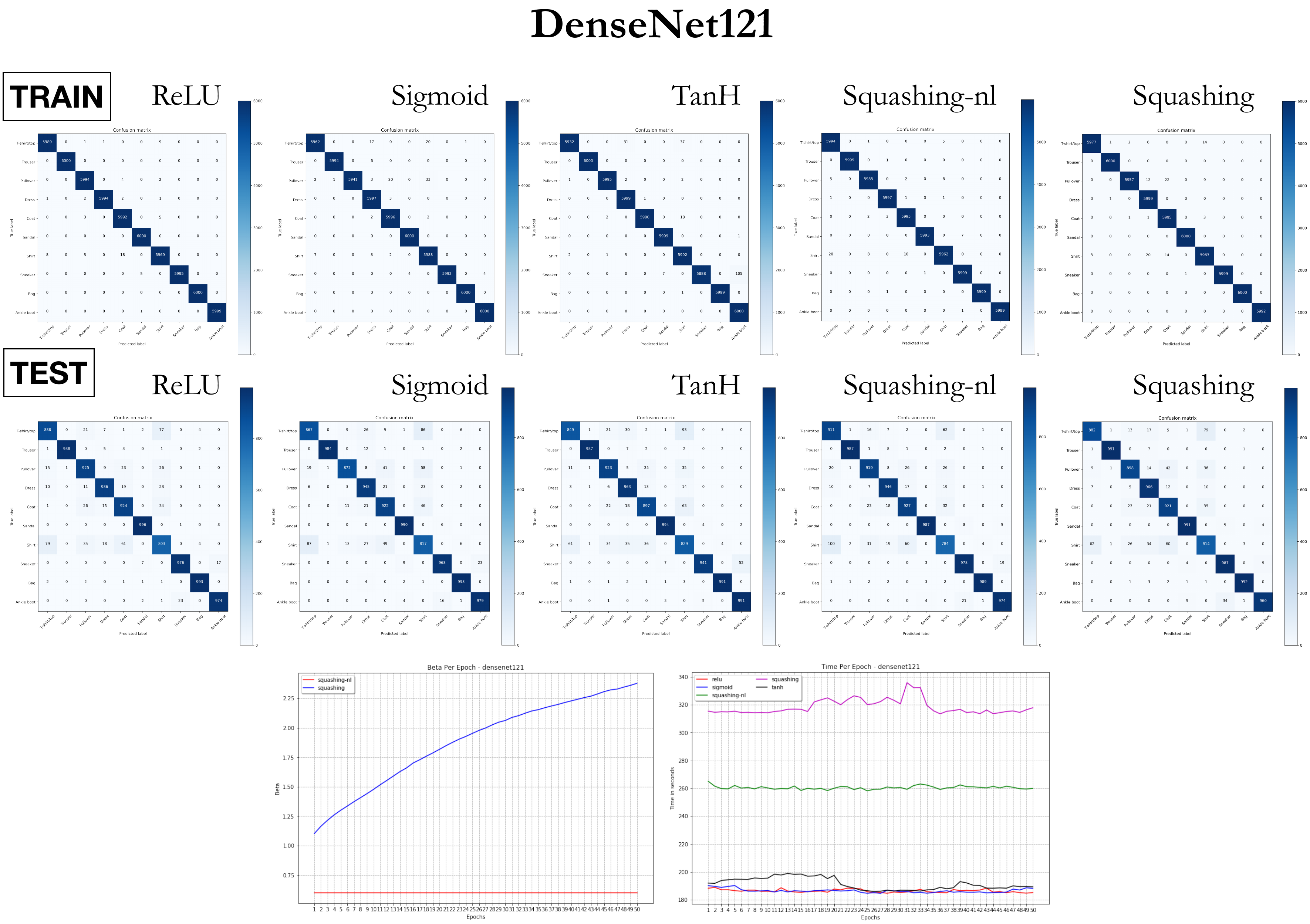}
		%	\caption{Line plot showing the time performance for different activation functions applied in the LeNet-5 architecture}
		%   \source{\footnotesize Own representation.}
		%	\label{fig:inception-v3_time}
	\end{figure}
\end{landscape}

\begin{landscape}
	\begin{figure}
		\centering
		\includegraphics[width=25cm]{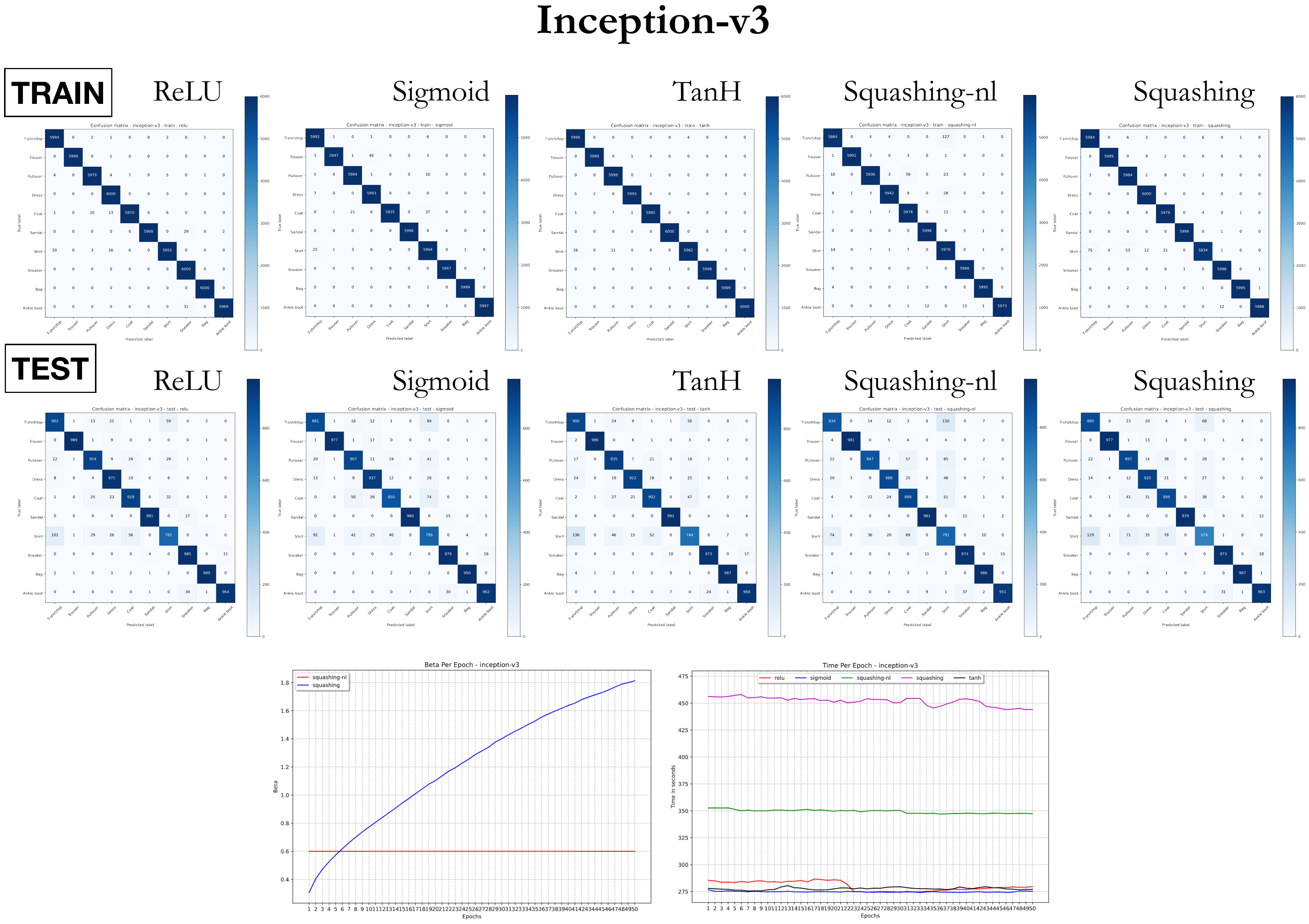}
		%	\caption{Line plot showing the time performance for different activation functions applied in the LeNet-5 architecture}
		%   \source{\footnotesize Own representation.}
		%	\label{fig:inception-v3_time}
	\end{figure}
\end{landscape}

\begin{landscape}
	\begin{figure}
		\centering
		\includegraphics[width=25cm]{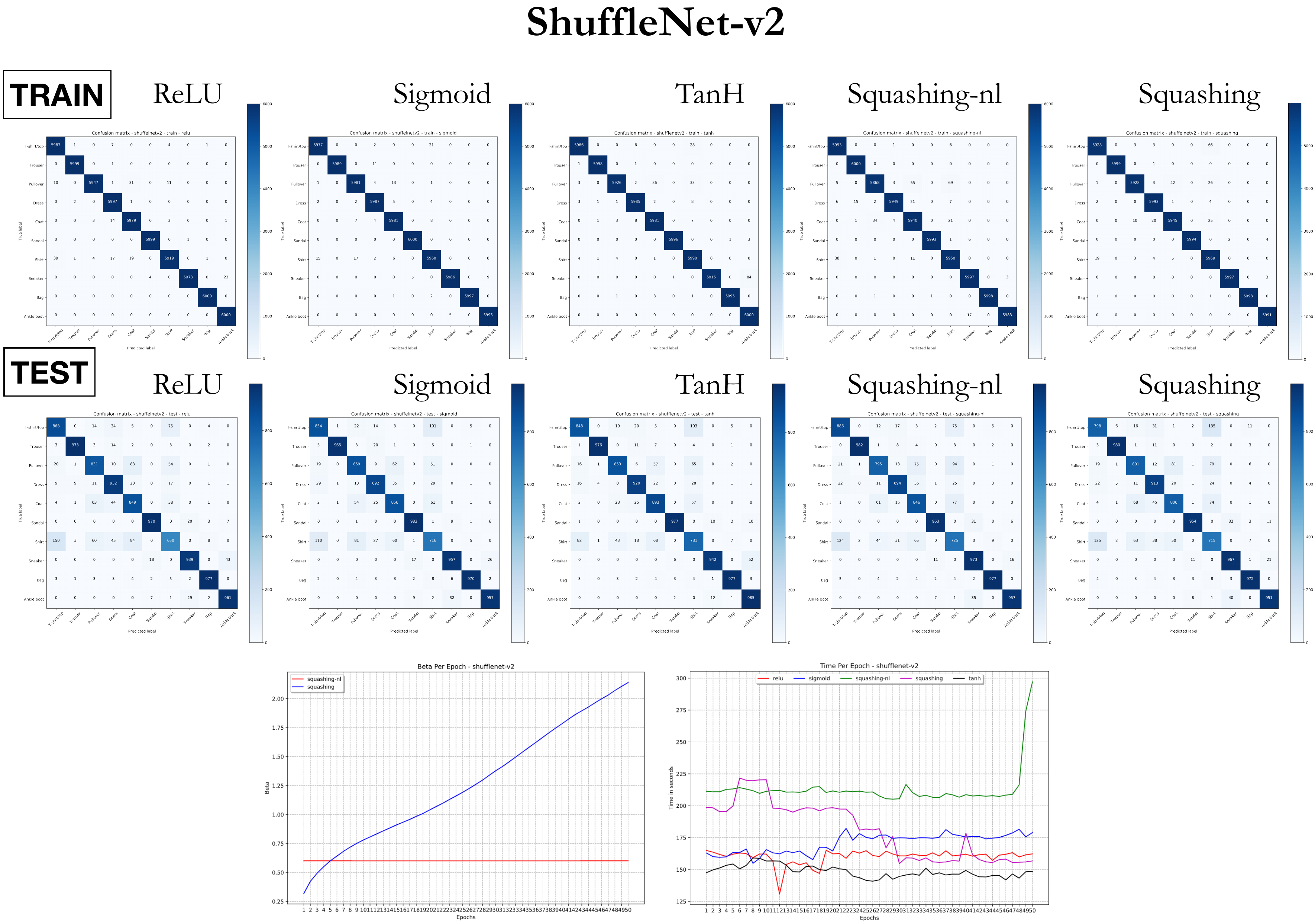}
		%	\caption{Line plot showing the time performance for different activation functions applied in the LeNet-5 architecture}
		%   \source{\footnotesize Own representation.}
		%	\label{fig:inception-v3_time}
	\end{figure}
\end{landscape}

\begin{landscape}
	\begin{figure}
		\centering
		\includegraphics[width=25cm]{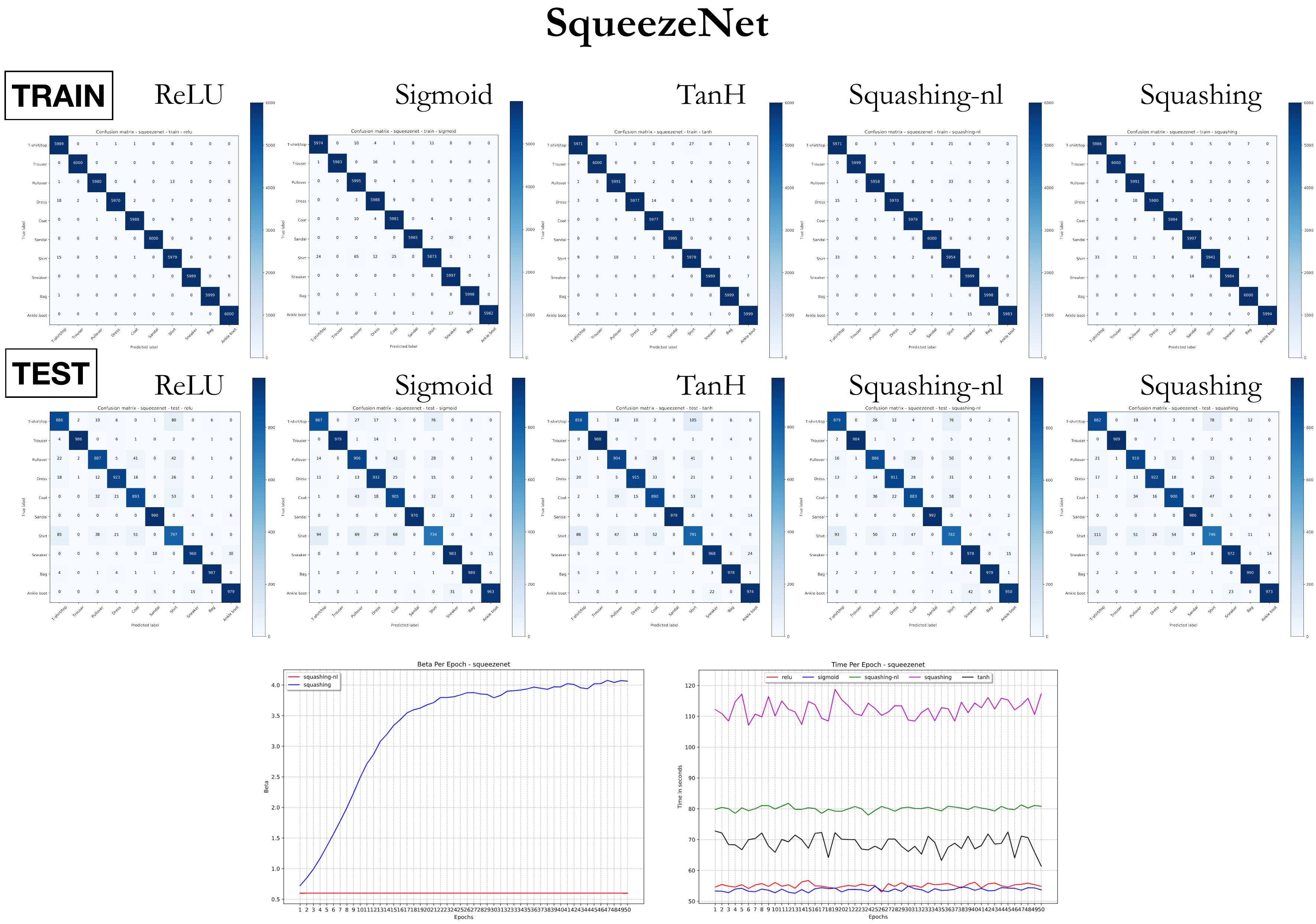}
		%	\caption{Line plot showing the time performance for different activation functions applied in the LeNet-5 architecture}
		%   \source{\footnotesize Own representation.}
		%	\label{fig:inception-v3_time}
	\end{figure}
\end{landscape}

\end{document}